%% file: main.tex
\pdfoutput=1

\documentclass[11pt]{article}

\usepackage{EMNLP2022}

\usepackage{times}
\usepackage{latexsym}

\usepackage[T1]{fontenc}

\usepackage[utf8]{inputenc}

\usepackage{microtype}

\usepackage{inconsolata}
\usepackage{amsmath}
\usepackage[shortlabels,inline]{enumitem}
\usepackage{cleveref}
\crefname{section}{s}{ss}
\crefname{section}{s}{ss}
\crefname{table}{Table}{}
\crefname{figure}{Fig.}{}
\crefname{algorithm}{Alg.}{}
\crefname{ALC@unique}{Line}{Lines}
\crefname{equation}{Eq.}{}
\crefname{appendix}{Appendix}{}
\crefformat{section}{\S#2#1#3}

\usepackage{todonotes}

\usepackage{multirow}
 
\title{POQue: Asking Participant-specific Outcome Questions for a Deeper Understanding of Complex Events}

\author{ %
\textbf{Sai Vallurupalli$^1$,
Sayontan Ghosh$^2$,} \\
\textbf{Katrin Erk$^3$, Niranjan Balasubramanian$^2$,
Francis Ferraro$^1$}
\\
$^1$ University of Maryland, Baltimore County, \\
$^2$ Stony Brook University, \\
$^3$ University of Texas, Austin \\
  \texttt{kolli@umbc.edu},
  \texttt{sagghosh@cs.stonybrook.edu},\\
  \texttt{katrin.erk@utexas.edu},
  \texttt{niranjan@cs.stonybrook.edu},
  \texttt{ferraro@umbc.edu}\\
  }
 
\begin{document}
\maketitle
\input{sections/abstract}
\input{sections/introduction}
\input{sections/related_work}

\input{sections/definitions}
\input{sections/dataset}

\input{sections/evaluation}

\input{sections/tasks}

\input{sections/models}
\input{sections/ablation_study}
\input{sections/conclusions}
\input{sections/limitations}

\section*{Acknowledgements}
We would like to thank the anonymous reviewers for their comments, questions, and suggestions. %
This material is based in part upon work supported by the National Science Foundation under Grant No. IIS-2024878. %
Some experiments were conducted on the UMBC HPCF, supported by the National Science Foundation under Grant No. CNS-1920079. %
This material is also based on research that is in part supported by the Army Research Laboratory, Grant No. W911NF2120076, and by the Air Force Research Laboratory (AFRL), DARPA, for the KAIROS program under agreement number FA8750-19-2-1003. The U.S.Government is authorized to reproduce and distribute reprints for Governmental purposes notwithstanding any copyright notation thereon. The views and conclusions contained herein are those of the authors and should not be interpreted as necessarily representing the official policies or endorsements, either express or implied, of the Air Force Research Laboratory (AFRL), DARPA, or the U.S. Government. %



\bibliography{anthology,custom}
\bibliographystyle{acl_natbib}

\newpage
\appendix

\input{sections/appendix-HIT}

\input{sections/appendix-processing}

\end{document}

%% file: sections/abstract.tex
\begin{abstract}
\label{sec:abstract}


Knowledge about outcomes is critical for complex event understanding but is hard to acquire. %
We show that by pre-identifying a participant in a complex event, crowdworkers are able to %
\begin{enumerate*}[(1)]
\item infer the collective impact of salient events that make up the situation,
\item annotate the volitional engagement of participants in causing the situation, and 
\item ground the outcome of the situation in state changes of the participants.
\end{enumerate*} %
By creating a multi-step interface and a careful quality control strategy, we collect a high quality annotated dataset of 8K short newswire narratives and ROCStories with high inter-annotator agreement (0.74-0.96 weighted Fleiss Kappa). %
Our dataset, POQue (\textbf{P}articipant \textbf{O}utcome \textbf{Que}stions), enables the exploration and development of models that address multiple aspects of semantic understanding. %
Experimentally, we show that current language models lag behind human performance in subtle ways through our task formulations that target abstract and specific comprehension of a complex event, its outcome, and a participant's influence over the event culmination.  
 
\end{abstract}

%% file: sections/introduction.tex
\section{Introduction}
\label{sec:introduction}



\begin{figure*}
\centering
\includegraphics[trim=0 .2in 0 0,  width=.95\linewidth]
{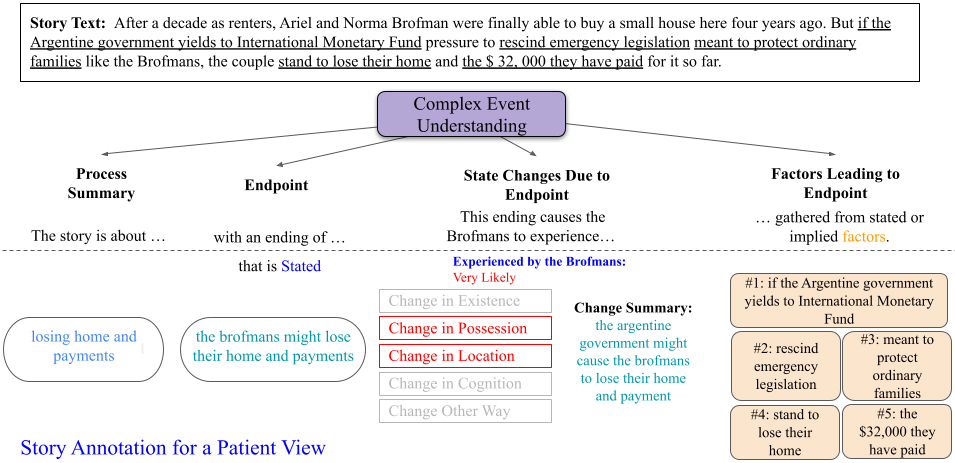}
\caption{Our approach to understanding state change outcome of complex events. Our four step annotation process involves describing abstractly what a story is about; writing an endpoint for the story; identifying and describing the changes that are a result of the endpoint, and identifying the salient sub-events that lead or support to those changes. To mitigate this complexity, we focus annotator's attention on one particular participant, and how that participant either causes or experiences the identified changes. This process has resulted in 4k annotated documents.} 
\label{fig1}
 \vspace{-4mm}
\end{figure*}

Situations that people experience or describe can be complex, and developing a computational understanding of these situations is not straightforward. %
Consider the short narrative from \cref{fig1}:
\begin{quote}
\small
After a decade as renters, [the Brofmans] were finally able to buy a small house here four years ago. But if the Argentine government yields to [IMF] pressure to rescind emergency legislation meant to protect ordinary families like the Brofmans, the couple stand to lose their home and the \$32,000 they have paid for it so far.
\end{quote}
Across multiple, interwoven events with multiple participants, this narrative describes part of the process of losing one's house. %
A \textit{possible} ending (the loss of a home) is suggested, which is the result of a confluence of these events. %
This ending can be semantically grounded in various changes of state that the participants experience, though note how the use of counterfactual considerations, conditional statements (``if the Argentine government...''), and varying levels of certainty over whether events have actually happened (e.g., realis vs. irrealis) contribute to the difficulty in understanding this complex event~\cite{Herman2002StoryLP,ryan_possible_worlds}. %

Knowledge about how event outcomes affect individual participants can help identify salient events in a narrative, fill in implicit missing
information~\cite{lobue-yates-2011-types} and chain events that lead to improved understanding of complex events~\cite{Graesser1994ConstructingID}.  To infuse AI models with similar knowledge, narrative comprehension research has focused on learning event relationships~\cite{mostafazadeh-etal-2016-caters,chambers-jurafsky-2008-unsupervised,ogorman_inproceedings,caselli-vossen-2016-storyline}, using temporal~\cite{Pustejovsky03timeml:robust}, causal~\cite{mirza-etal-2014-annotating} and discourse~\cite{prasad_inproceedings} relationships in text. %
However, as \citet{dunietz-etal-2020-test} and \citet{piper-etal-2021-narrative} argue, for a more useful, generalizable, and robust comprehension, we need to take a holistic view of complex events. %
In this paper, we tackle an understudied notion of this holistic view and examine knowledge of post-conditions based in states to support inferences of the form ``who did what to whom and with what end result.''

A core insight we make is that viewing complex events through the lens of a single participant at a time, either from an agent (how the participant affects others) or a patient (how the participant is affected) view, can help mitigate the complexities we have discussed. %
In this we build on cross-disciplinary research that shows that humans mentally structure events along single participants~\cite{Black1980STORYUA,MORROW1989292}, and that participant-based event and outcome analysis improves complex event understanding~\citep{Dijk1983StrategiesOD,liveley2019narratology}. 
We also note that a complex event is not exhaustively described by what is stated in the text: it is well known that speakers often omit narrative steps that can be inferred~\cite{Grice1975}, including outcomes and effects of a narrative that are often left implicit.  

We present POQue, a dataset with post-conditional knowledge about complex events. We identify cumulative outcome-oriented endpoints of the stories caused by related events and the consequences or post-conditions of those events as state-based changes in participants.  Seen in \cref{fig1}, in a storyline involving a participant ("the Brofmans"), we identify an ending outcome for the complex event (the \textit{Endpoint}, "the brofmans might lose their home and payments"), with salient events that lead to this (\textit{Factors Leading to the Endpoint}).  We relate a participant's involvement in the complex event (as a "patient" who "Very Likely" experienced the ending outcome) and the changes of state occurring as a result of the complex event (the changes in possession and location experienced by the Brofmans and other families, and the change in possession by the Argentine government and IMF).      

To facilitate high quality annotations we designed a multi-stage crowd sourcing solution to acquire, monitor, assess and curate annotations at scale.  We collected 7772 annotations across 4001 stories and assessed a random 1545 annotations (20\%) in a multistage pipeline to obtain a highly curated test set. %
Using POQue, we test current language models on reasoning about complex events in narratives: we formulate challenge tasks to identify and generate post-conditions from a story, and evaluate how well trained models predict a participant's involvement in enabling a complex event.

We summarize our contributions as follows: 
\begin{enumerate*}[(1)]
\item we introduce a new annotation scheme focusing on complex events from the point of view of a single participant.
\item We create a new dataset of complex events from three collections of everyday stories, using free form text to obtain insight into implied outcomes. 
\item We obtain high quality annotations from crowd workers without the use of requester generated qualification tests. 
\item We formulate challenge tasks aimed at evaluating the ability of language models to perform richer complex event comprehension, specifically: 
a) generating a process summary of the complex event b) generating an endpoint of the complex event, c) generating the outcome of a complex event based on a participant's semantic role d) identifying a participant's involvement in a complex event, and e) generating post-conditions or changes caused by a complex event. 
\end{enumerate*}
Our dataset and code are publicly available at \href{https://github.com/saiumbc/POQue}{https://github.com/saiumbc/POQue}. %

%% file: sections/related_work.tex
\section{Related Work}
\label{sec:related_work}

Narrative texts communicate experiences and situations by connecting related events \cite{brooks2012reading,Mateas_NI} through events involving participants \cite{bal1997narratology,eisenberg-finlayson-2017-simpler,liveley2019narratology}. %
Previous works, viewing narratives as sequences of events, annotated event pairs for event coreference, temporal, and causal relationships~\cite{ogorman_inproceedings,caselli-vossen-2016-storyline,mirza-tonelli-2016-catena,mostafazadeh-etal-2016-caters}.
Newer works have studied event groups using predicate hierarchies~\cite{qi-etal-2022-capturing} and temporal graphs~\cite{li-etal-2021-future}.  However, these approaches focus on event-event relationships, without diving deeply into participant or entity analysis. %
Unsupervised methods assume narratives are coherent and learn partially ordered event chains  \cite{chambers-jurafsky-2008-unsupervised,balasubramanian-etal-2013-generating} or sub event relationships \cite{yao-etal-2020-weakly} but these are limited to what occurs in the text itself, which can lead to well-known issues of bias or evaluation limitations~\cite{gordon2013reporting,rudinger-EtAl:2015:EMNLP}.

\newcite{caselli-inel-2018-crowdsourcing} obtain crowd annotations of causal relationships between events and assess their quality by relating them to expert annotations. %
PeKo~\cite{kwon-etal-2020-modeling} uses crowd annotations of precondition relationships and fine tunes a language model for finding such relationships. %
However, both works limit their study to event pairs in short text snippets. %
The ESTER dataset~\cite{han2021ester} consists of more comprehensive relationships in a story, though limited to within-text (i.e., stated) mentions of events only. 
GLUCOSE~\cite{mostafazadeh-etal-2020-glucose} provides elaborate causal relationships for several event dimensions for each event in a story. 
However, in contrast to our effort, these works address direct causal relationships between within-text events and do not focus on participants.

Understanding complex events has long attracted cross-disciplinary attention. %
For example, theoretical linguistics and cognitive science work has shown that humans understand a narrative text using simulative inference~\cite{kaplan_schubert,Boella_NU,schubert_episodiclogic}. %
Prior work has also shown how observing participants' events and the resulting consequences can lead to improved understanding of events~\cite{Dijk1983StrategiesOD,zwaan}.  


%% file: sections/definitions.tex
\section{Knowledge Representation}
\label{sec:Definitions}

As an underlying motivation for our efforts, we posit that for language models to be able to reason about complex events from narratives, they should be able to identify a likely ending of that complex event, component events that lead to that ending, and the state changes that result from that ending. %
However, this type of knowledge is complex and has been computationally understudied, leading to a scarcity of sizable datasets. %
In our efforts to correct this, we appeal to classic, cognitively- and linguistically-backed results. %

First, inspired by the idea from \citet{Kintsch1978TowardAM} that text comprehension involves reducing relevant details into an abstract coherent semantic text, our targeted annotations include an abstract high level summary and the minimal set of salient events that make up the complex event. %
Second, we extend the idea of thematic roles for verbal arguments~\cite{dowty} to a generalized semantic role for the complex event. %
Specifically, Dowty showed that easily verifiable characteristics and properties, such as volitional participation in an event or whether a participant underwent a change of state because of a \textit{particular} event, can be used to define predicate-level prototypical semantic roles. %
Inspired by this, we characterize the roles of complex events through the intentional engagement in changes of state of participants. %
As such, our targeted annotations account for both the intentional involvement of the participants in the complex event and the cumulative impact of all the events that make up the complex event. %


\paragraph{Story and Participant} 
Stories in our dataset are either a ROCStory or heuristically salient portions of newswire (first 100-150 tokens).\footnote{While we acknowledge they have important differences, we use ``narrative'' and ``story'' interchangeably.}  For more information on story processing see \cref{sec:story-prep}. %
We define a participant as an entity that was mentioned several times in the story.  See \cref{sec:story-prep} for more on participant selection.  Multiple entities are considered a single Participant if these entities are mentioned together and participate together in all the events. In \cref{fig1}, "the Brofmans" are a Participant.

\subsection{Targeted Knowledge Annotations}

Given a story $S$, a participant $P$, and $P$ taking on a agent-like or patient-like cumulative semantic role, $PR$, we obtain the following annotations. 


\paragraph{Process Summary (PS):} A high-level, free-form description of the situation, which provides the topical context for the complex event. 
For example, ``Losing home and payments'' for the story in \cref{fig1}.

\paragraph{Endpoint Description (ED):} A free-form description of the inference of what happened or is likely to happen in the story, conditioned on the process summary.  
It is the result of an aggregate of story events that leads to state changes for participants.  
For the story in \cref{fig1}, the endpoint is ``The Brofmans might lose their home and payments.''

\paragraph{Endpoint Anchoring (EA):} An endpoint may be (inferentially) \textit{stated} in the text, or it may be \textit{implied}/suggested. 
We judge this via a three-way choice (stated, implied, unsure). %
For the story in \cref{fig1}, the endpoint is \textit{stated} by the text.

\paragraph{Participant Involvement (PI):} Whether the participant caused the endpoint, or experienced it, indicated with a 5 point Likert scale, from very unlikely to have caused [experienced] the endpoint, to very likely to have caused [experienced] it. For the story in \cref{fig1}, 
the complex event maximally affects $P$.  Hence this rating is a ``very likely.''  
 
\paragraph{Change Summary (CS):} A templated text description of state changes caused or experienced by participants as a result of the endpoint.  In \cref{fig1}, this is ``The Argentine government might cause the Brofmans to lose their home and payment.''

\paragraph{Change Modes ($\bf(c_1, c_2,..,c_k)$):} The various ways in which participants experience changes.  These change modes are: change in existence, feeling, location, possession, some other way, or no changes. This list was inspired by classic linguistics, c.f., \citet{dowty}, though refined during early examination of our stories.  

\paragraph{Factors ($\bf(f_1, f_2, .., f_n)$):}
Salient events that lead to the endpoint and state changes where each factor captures an event in a phrase comprising of at least a subject and verb. For the story in \cref{fig1}, ```if the Argentine government yields to International Monetary Fund,'' 
is one of the factors leading to the Endpoint. %

\subsection{Crowd Annotations}
We created a human intelligence task (HIT), deployed on Amazon Mechanical Turk (AMT). 
The HIT consists of a story with highlighted participant mentions displayed in the left column and four annotation steps in the right column which vary slightly for the agent and patient views. The protocol was IRB approved. 

Crowd workers were instructed to read the story, focus on the highlighted participant, and provide annotations. 
We provide several annotated examples, general instructions for completing the HIT, and specific instructions for each step suggesting a template to follow for some steps. More details and the layout of the HIT are in \cref{app:HIT_information}.

The annotation task consists of 4 steps and each story is assigned to two workers, one where the highlighted participant is assigned the role of ``agent'' and another with the assigned role as ``patient.'' Step 1 asks for a high level description of the story, a process summary of the situation described.  Step 2 of the HIT asks for a description of an endpoint in the story.  We assume a story's endpoint typically signifies a state change caused by a complex event. %
Step 3 asks for a summary of changes caused by the complex event in the story participants and also asks to identify the type of changes. 
Step 4 of the HIT asks an annotator to identify the salient events, or factors, that lead up to the complex event and changes from it.   

%% file: sections/dataset.tex
\section{Dataset}
\label{sec:dataset}

We selected stories from three narrative English language datasets -- the ESTER dataset~\cite{han2021ester}, the ROC stories dataset~\cite{mostafazadeh-etal-2016-caters} and passages from the Annotated New York Times newswire dataset~\cite{Sandhaus2008Nyt}. We selected these given the prominence the underlying documents have in the broader NLP community (the ESTER  documents are a subset of the TempEval3 (TE3) workshop dataset~\cite{uzzaman-etal-2013-semeval}. We included ROC stories because they often contain a single situation with mostly salient information. We noticed these stories help crowd workers easily focus on salient events, providing cues for factors and state changes. Meanwhile, ESTER and the selected Newswire stories provide a variety of complex situations and discourse text. %
Additionally, by selecting subsets of these well-known datasets, we hope that future efforts may be able to 
aggregate our annotations with existing ones, enabling richer phenomena to be examined.

\subsection{Story Preparation}
\label{sec:story-prep}
We sampled stories from the Annotated New York Times (ANYT) corpus, ROCStories, and ESTER. We then identified participants via an automatic entity coreference system. We heuristically selected relevant and annotable excerpts of the document by identifying ``continuant story lines'' (see \cref{app:dataset:prep}). After identifying a participant and a continuant story line, we randomly selected 4001 stories for annotation; see \cref{tab1} for details. 

\subsection{Dataset Annotation \& Pricing}
\label{sec:data-annotation}
Our dataset has 4001 stories, annotated by 163 different crowd workers. The average number of stories annotated per worker is 43. When possible, we annotated from both an agent and a patient view for a participant, so in total we obtained 7773 annotations. %
Workers were paid an average of \$0.50 for annotating a single HIT, either an agent or a patient view of the story.  For more detailed information on payment and training see \cref{app:HIT_information:training-pay}.
We tackle the positivity bias in AMT work~\cite{mturk_article} using a thorough initial verification and training (see \cref{app:HIT_information}) and ensured workers understood the task and provided quality annotations.



\begin{table}
\centering
\begin{tabular}{|l|r|r|r|}
\hline
  &  \emph{\# Stories} & \emph{\# Agent} & \emph{\# Patient} \\
\hline
Total  & 4001 & 3896 & 3876\\
ROC  &  1383 & 1364 & 1356\\
ESTER  & 1275 & 1237 & 1218 \\
NYT  & 1343 & 1295 & 1302\\
\hline

\end{tabular}
\caption{Document-level data statistics. Note that the number of stories refers to the number of unique stories annotated, while the agent and patient numbers refer to the number of instances annotated on those documents. Additionally, 260 of the ROCStories are from the CATERS~\cite{mostafazadeh-etal-2016-caters} collection.  CATERS stories and ESTER stories containing subevents are useful for relating causal and compositional events in a complex event. }
\label{tab1}
\vspace{-2mm}
\end{table}

\begin{table}[t]
\centering
\small
\begin{tabular}[trim=0 .2in 0 0,  width=.95\linewidth]{p{0.6\linewidth} | p{0.3\linewidth}}
\hline\hline
Avg. process summary length & 5.7 words \\ \hline
Avg. endpoint length & 9.4 words \\ \hline
Endpoint stated/implied/unsure \% & 68.5\%/28.9\%/2.6\% \\ \hline
Avg. change descr. length & 8.9 words \\ \hline
Avg. likelihood of causing change (agent) & 4.0 (likely) \\ \hline
Avg. likelihood of experiencing change (patient) & 4.1 (likely) \\  \hline
Avg. \# of factors & 3.7 \\ \hline
Avg. factor length & 8.0 words \\ \hline
\end{tabular}
\caption{Additional statistics about POQue.}
\label{tab:annotation-stats}
\vspace{-4mm}
\end{table}

\begin{figure}[t]
\includegraphics[trim=0 .2in 0 0,  width=.95\linewidth]
{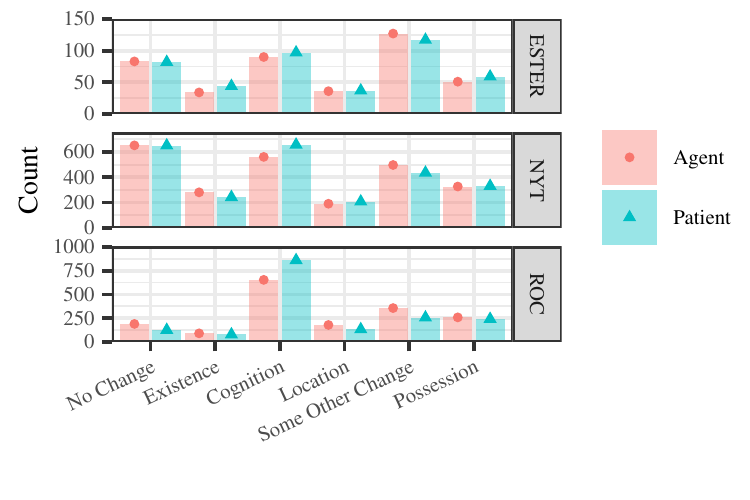}
\vspace{-1em}
\caption{Count of change modes, shown for each of the agent and patient roles, and broken out across the originating datasets our annotated narratives come from.}
\label{fig:count-change-modes}
\vspace{-4mm}
\end{figure}

\subsection{Dataset Statistics}

For most stories we obtained two annotations, with the two participant semantic roles. %
We show high-level document statistics in \cref{tab1}. Annotations for the two different roles of the highlighted participant are shown in separate columns. %
We show more detailed annotation statistics in \cref{tab:annotation-stats}, and examine the frequency of change modes in \cref{fig:count-change-modes}. %

%% file: sections/evaluation.tex
\subsection{Dataset Quality and  Analysis}
\label{sec:evaluation}

We noticed that the nature of the stories and the task steps elicit a variation in the text style and format, even from the same worker.  Our experiments (\cref{sec:models}) and ablation studies (\cref{sec:ablation-study}) did not uncover any easy biases attributable to a small number of workers producing most of the annotations.  Due to space limitations we explain our process for evaluating 1545 random annotations in \cref{app:ValidationHITs}.

%% file: sections/Tasks.tex
\section{Tasks for State Change Knowledge}
\label{sec:tasks}

Based upon our collected dataset, we propose several tasks. These tasks are designed to test various aspects of comprehension involving complex events, their participants, and outcomes.

\subsection{Task 1: Generating Process Summaries}

Categorizing stories based on the type of situation they describe is necessary for generalization.  For this, we fine-tune models to generate an abstract and high level process summary of the complex action described in a story.  Because we annotated salient events for the story, i.e., the factors, we have two task formulations.  We fine-tune models to generate $PS$ either given $S$ or $(f_1, f_2,..,f_n)$. Both are standard summarization tasks which we compare with a baseline where the process summary is assumed to be ``About $P$.''

\subsection{Task 2: Generating Complex Event Endpoints and Salient Events}

Understanding a story involves the identification and decomposition of salient events that lead to an endpoint, for the described complex event.  We test this understanding with two complementary formulations where we generate either the endpoint description or the salient events, i.e., factors.   For generating $ED$ we have two sub formulations where we fine-tune models either on $S$ or $(f_1, f_2,..,f_n)$.  For generating $(f_1, f_2,..,f_n)$  we fine-tune models on $(S, ED)$. These are all standard summarization tasks which we compare with a baseline where $ED$ is assumed to be the last sentence of the story.



\subsection{Task 3: Generating Changes Resulting from a Complex Event}

Knowing the changes caused by a complex event gives us an insight into its importance and the intentions (addressed in Task 5) behind it.  In this task, we generate changes caused by a complex event through the lens of the semantic role tracking we have employed throughout our effort.  Using standard summarization, we fine-tune models to generate $CS$ given $(S, ED, PR)$. %

\subsection{Task 4: Identifying Types of Changes}
Grounding the impact of a complex event in the various change modes a participant undergoes helps in understanding the importance of new situations by relating them to known situations with similar post-conditions.  
We formulate this as a multi-label binary classification and fine-tune models to identify $k=5$ change modes  $(c_1, c_2,..,c_k)$ given $S$.

\subsection{Task 5: Assessing Participant's Involvement in the Complex Event}
Besides the story context, the participant's semantic role heavily influences our decision of whether the participant intended or enabled the complex event or the changes caused by it. In this binary classification task, we predict the participant involvement rating $PI$ given $(S, ED, PR)$.  This prediction demonstrates a model's ability to identify a participant's intentional engagement and enablement of the complex event and its impact.


%% file: sections/models.tex
\section{State Change Benchmark Experiments}
\label{sec:models}
We benchmark the performance of current encoder-decoder transformer language models, T5~\cite{JMLR:v21:20-074} and BART~\cite{lewis-etal-2020-bart}, which are effective for both text generation and classification.  We compare fine-tuned base and large\footnote{We found that training a bart-large model was finicky, with some training runs not converging. For these cases, we do not include  results for the bart-large model. See https://github.com/huggingface/transformers/issues/15559. 
} models with multiple automated metrics and crowd sourced human evaluation.  We use bootstrap for calculating statistical significance via the \texttt{mlxtend} library~\citep{raschkas_2018_mlxtend}.  



\subsection{Automated Evaluation}
We use the classic metrics of  \textbf{ROUGE}~\cite{lin-2004-rouge}, \textbf{BLEU}~\cite{BLEU}, and \textbf{METEOR}~\cite{lavie-agarwal-2007-meteor}, and the more recent \textbf{BertScore}~\cite{bert-score}. %
Due to space limitations, we present ROUGE-L and BertScore in the main paper, and additional ROUGE-1, ROUGE-2, METEOR, and BLEU scores in the appendix (\cref{sec:expanded-results}).
We use standard metrics used for single and multi-label classification: \textbf{Accuracy} and \textbf{macro F1}. In multi-label classification, we calcuate \textbf{Subset Accuracy} and \textbf{macro F1} using sklearn and a \textbf{Hamming Score} which is computed as $\frac{1}{n}\Sigma_{i=1}^{n}\frac{Y_i\cap Z_i}{Y_i\cup Z_i}$, where $Y$ and $Z$ are true and predicted labels for $n$ examples.


\subsection{Human Study of Model Generations}
\label{sec:human-eval-generation-description}
We perform a human evaluation of the generation tasks (1, 2, and 3) using 50 randomly selected generations for each model and the corresponding human annotations.  We obtained qualitative ratings from 3 crowd workers experienced in annotating our HITs and  measured IAA using a weighted Fleiss's Kappa as in \cref{app:annotation-quality-assessment}. For each summary, workers are presented with the story and the summary and asked to rate the summary on aspects that relate to the task such as abstractness, factuality and salience using a 5-point Likert scale. See \cref{app:human-evaluation-HITs} for more information on these aspects and the HITs 
used for evaluation. 

\subsection{Task 1: Generating Process Summaries}


To test whether a model generates a more focused process summary when trained on salient information, we compare pre-trained models fine-tuned on $S$ and $(f_1, f_2,..,f_n)$ with an easy baseline process summary of ``About $P$,'' where $P$ is the participant. Less than 1\% of the process summaries in dataset and model generations contain this baseline format. 
Results from this task training are listed partially in \cref{tab:generating-actions} and fully in \cref{app:generating-actions}. For all models, Rouge, BLEU and METEOR scores show less lexical overlap, but BertScore indicates a high similarity between the model generated and reference summaries.   Inspired by previous work in measuring abstractiveness~\cite{DBLP:journals/corr/SeeLM17,DBLP:journals/corr/abs-2108-02859,gao-etal-2019-write,narayan-etal-2018-dont}, we compare average number of tokens (\textit{Len}) across all summaries, the  percentage of exactly matched trigram spans in the story (\textit{Ext}), and the average of Abstractness Likert scores (\textit{Abstr.}) from the evaluation HIT (see \cref{app:evaluation_process_summary}) for process summaries. %
We provide Len, Ext and Abstr. values to help contextualize scores. While a lower value of Ext does not necessarily mean a better generation, it does mean there is less direct copying of length-3 phrases.

\begin{table}[t]
\centering
\resizebox{.98\columnwidth}{!}{
\begin{tabular}{c|l|c|c||c|c||c|}
\cline{2-7}
 & Model & Len & Ext. ($\downarrow$) & RougeL & BertScr & Abstr. \\
\cline{2-7}
& Reference &\textbf{3.6}&\textbf{.13*}&\multicolumn{2}{|c||}{-}& \textbf{3.57*} \\  \cline{2-7}
& About $P$&1.7&.27&10.43& 83.86& 2.37   \\
\hline
\hline
\multirow{3}{*}{Story} & Bart-base &4.0&.46& 21.43 & 86.70 & 2.77  \\
& T5-base  &10.0&.60& 19.50& 85.99 & 2.32   \\
& T5-large &6.9&.63& 20.30 & 86.15 & 2.13   \\
\hline
 \hline
\multirow{3}{*}{Fact.} & Bart-base &4.2&.33& \textbf{23.81} & \textbf{87.66} & 3.22   \\
 & T5-base &9.9&.56& 18.29 & 86.10 & 2.73   \\
\cline{2-7}
\end{tabular}
}
\caption{Generating Process Summaries (Task 1).  See appendix \cref{app:generating-actions} for the full results. Bart-large is not included because we were unable to get it to properly converge. The best scores are bolded. We use * to indicate a significantly higher value than other values in the column with a p value between 0.001 and 0.0001 (except for Bart-base trained on Factors where the p value is 0.13). \textit{Len} value for Reference is considered the best as the baseline value is not meaningful. }   
 \label{tab:generating-actions}
 \vspace{-5mm}
\end{table}

\noindent\textbf{Discussion:} BART generations are brief, less extractive and more abstractive, whereas T5 generations are longer, less abstractive and more directly drawing upon spans of story text. The Reference summaries are brief, significantly less extractive and at a significantly higher abstractness score compared to all the models. Models fine-tuned only on the factors produce 
more abstractive summaries. However, this increase in the abstractness for the BART model increased the factual errors, in line with previous observations~\cite{DBLP:journals/corr/abs-1711-04434,kryscinski-etal-2019-neural,DBLP:journals/corr/abs-2108-02859}.    The significantly higher brevity and abstractness of the Reference summaries point to a substantial gap between human and LMs' ability at capturing complex actions in a brief, high-level, abstract phrase.

\subsection{Task 2: Generating Endpoints \& Factors}

To test how well models generate endpoints, models are fine-tuned to generate $ED$, given  $S$ or $(f_1, f_2,..,f_n)$ and compared with the baseline version where the $ED$ is assumed to be the story's last sentence.   Partial results for this task formulation are listed in \cref{tab:gen-action-events} and the full results in \cref{app:gen-action-events} for the trained models. We also compare models trained on the complementary task of generating $(f_1, f_2,..,f_n)$ given ($S$, $ED$).  A special token separates factors in all the task formulations involving factors.  Partial results for this complementary task are listed in \cref{tab:gen-action-factors} with the full results in \cref{app:gen-action-factors}.

\begin{table}[t]
 \centering
 \resizebox{.98\columnwidth}{!}{
 \begin{tabular}{l|l|c|c|c|c||c|c|}
 \cline{2-8}
 &  Model & Len & Ext$\downarrow$ & RougeL &   BertScr & Fact. & Sal.\\ 
  \cline{2-8}
  & Reference &7.9 &.27 & \multicolumn{2}{|c||}{-}&4.15  &3.46  \\ 
 \cline{2-8}
 & Last sent. & 23.3 & .79 & 21.82 & 85.60 & 4.49 & 3.35  \\
\cline{2-8}
\hline
 \multirow{4}{*}{Story} & Bart-base  & 11 & .72 & 25.43 & 87.61 & 4.66 & 3.97\\
& Bart-large & 10.3 & .63 & 24.74 & 87.62 & 4.59 &  3.81  \\
& T5-base  & 13.6 &.70 &  24.07 & 87.19 & \textbf{4.71*} & 4.03   \\
& T5-large & 12.9 &.67 & \textbf{25.71} &  87.54 & \textbf{4.71*}& \textbf{4.23*} \\
 \hline
 \hline
\multirow{2}{*}{Fact.} & Bart-base & 7.3 & .49 & 24.09 & \textbf{87.93} &  4.11 & 3.28 \\
& T5-base  & 10.4 & .47 & 22.01 &  87.07 & 3.99 & 3.02  \\
 \cline{2-8}
 \end{tabular}
 }
 \caption{Generating Endpoints for stories and factors (Task 2a). See appendix \cref{app:gen-action-events} for the full results.  The best scores are bolded. * indicates the value is significantly higher than the Reference value with a p value of .002 for Factuality and .0008 for Salience.   
 }
\label{tab:gen-action-events}
\vspace{-2mm}
\end{table}

\begin{table}[t]
 \centering
\resizebox{.98\columnwidth}{!}{ \begin{tabular}{|l|c|c|c|c||c|c|c|}
 \hline
 Model  & \# fact. & Len & RougeL & BertS & Brev. & Fact. & Sal. \\
  \hline
 Reference &3.6 &8.3 &\multicolumn{2}{|c||}{-}&\textbf{3.35*}  &3.25&3.04 \\ 
\hline
\hline
 Bart-base  & 3.5 & 14.0 &  45.28 & 88.06 & 2.54 & 3.49 &  3.57   \\
 Bart-large  & 3.6 &13.7 &  45.98 &  88.10 &  2.12  & 3.2  & 3.31 \\
T5-base    & 2.6 & 19.6 & 43.31 & 87.74 &  3.23 & 3.69 & \textbf{4.01*}  \\
 T5-large   & 3.7 & 13.9 & \textbf{47.96} & \textbf{88.44} & 2.85  & \textbf{3.80*} & 3.96*  \\
 \hline
 \end{tabular}
 }
\caption{Generating Factors from stories and their endpoints (Task 2b).  See appendix \cref{app:gen-action-factors} for the full results.  
The best scores are bolded.  *  indicates the value is significantly higher than the Reference value for Factuality and Salience with a p value of .0001. The Reference value for Brevity is significantly higher than all values in the column with a p value of .0001. 
}
\label{tab:gen-action-factors}
 \vspace{-4mm}
\end{table}

For all models, Rouge, BLEU and METEOR scores show higher lexical overlap, and BertScore indicates a high similarity between the generated and reference summaries.  We compare the average of the Factuality and Salience scores (\textit{Fact.} and \textit{Sal.}, resp.) from the endpoint summary evaluation HIT (see \cref{app:evaluation_endpoint_summary}) along with the average number of tokens (\textit{Len}) across all summaries and the  percentage of exactly matched trigram spans in the story (\textit{Ext}).  We also compare the average of the Brevity, Factuality and Salience scores (\textit{Brev.}, \textit{Fact.}, and \textit{Sal.}) from the evaluation HIT (see \cref{app:evaluation_factors}) for factors along with the average number of factors (\textit{\# fact.}) across all stories and the average number of tokens (\textit{Len}) across all factors and stories. 

\begin{figure}[t]
\includegraphics[trim=.8mm 0 0 0,  width=.95\linewidth]
{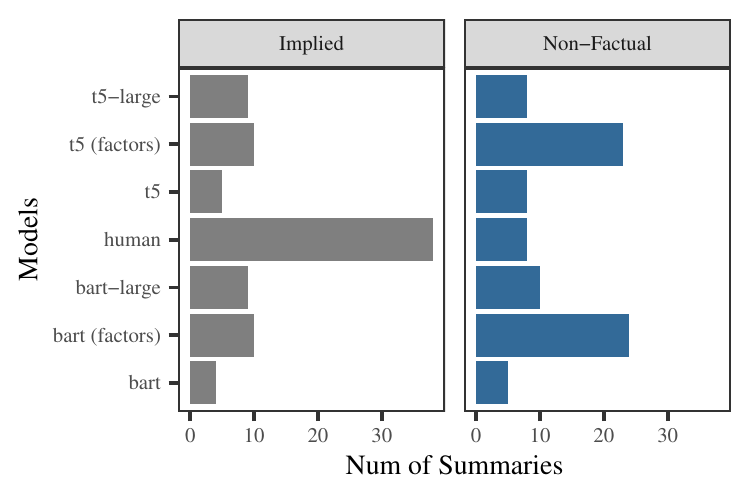}
\vspace{-1em}
\caption{Non-factual and implied endpoint types.}
\label{fig:endpoint-types}
\vspace{-4mm}
\end{figure}

\noindent\textbf{Discussion:}  Scores are significantly higher for LM generations than Reference endpoint descriptions on both Factuality and Salience.  We looked at a random 100 Reference endpoint descriptions and the corresponding model generations. The first author of this paper identified which of the endpoints were not directly stated in the story, but rather implied by the story, and which ones were not-factual.  
As shown in \cref{fig:endpoint-types}, very few of the model generations are implied endpoints and a third of the Reference endpoints contain implied  descriptions.  Our HIT instructions asked workers to annotate not only explicit endpoints but also the ones implied by the story and they identified 29\% of them as implied.  Evaluators lowered their scores both for factuality and salience for these implied endpoints as the description may be a possible but not strictly entailed outcome.  Models trained on factors generated more implied endpoint descriptions but  these implied endpoints contained more factual errors possibly because of less available context. We conclude that LMs try to generate stated endpoint descriptions unless challenged by limited context. 


Fine-tuning on stories and endpoints generates salient factors, indicated by the high assessment scores. However, the generated factors on average contain multiple facts making them less concise and focused than human written factors.

\subsection{Task 3: Generating Change Summaries}

We compare model generations of state changes, by fine-tuning models to generate $CS$ conditioned on $(ED,PR)$.   Given the pair $(S, t)$, where
$t$ is either 
``$P$~caused~this:~$ED$'' if $PR$=``agent'' or  ``$P$~experienced~this:~$ED$'' if $PR$=``patient'', fine-tuned models generate $CS$.
Partial results for this task are listed in \cref{tab:gen-changes-from-action}, and the full results in \cref{app:gen-changes-from-action}.  For all models, Rouge, BLEU and METEOR scores show high lexical overlap and BertScore indicates a high similarity between model generations and Reference summaries. We compare the average of the Factuality and Salience Likert scores (\textit{Fact.} and \textit{Sal.} resp.) from the  evaluation HIT (see \cref{app:evaluation_change_summary}) which measures whether the generated text contains change(s) resulting from the complex action.   

\noindent\textbf{Discussion:} 
T5 models generate change summaries that are significantly more factual and salient than the BART models.  While T5 generations score higher than Reference summaries, the difference is not significant.  To explain these results, we inspected the 50 evaluated stories and found that less than 10\% of the stories have no changes even though annotators indicated ``no changes'' for 25\% of the stories in these 50 (and in the entire dataset).  To see if crowd workers can identify these no-change stories, we ran a HIT where the change summary for these 50 stories was set to no-changes. From the results of this HIT we found that human evaluators also think there are 3 times as many stories with no-changes.  Workers missed subtle changes in a story especially when they relate to changes in cognition.  T5  was able to identify story text that contained subtle changes while the BART models seem to be learning the data distribution from the training data.  BART generations also contain a higher number of factual errors leading to its subpar performance.   %

\begin{table}[t]
\centering
\resizebox{.98\columnwidth}{!}{ \begin{tabular}{|l|c|c|c|c|}
 \hline
 Model  & RougeL & BertScr & Fact. & Sal.\\

 \hline
  Reference & \multicolumn{2}{|c|}{-} & 3.36 & 3.32 \\   \hline
\hline
Bart-base  &  \textbf{34.79} &  \textbf{88.39} & 3.03 & 2.93 \\
 Bart-large &  32.80 &  88.23 &  2.99 &3.05\\
 T5-base  &   26.81 &  87.20  & 3.74 & 3.23\\
 T5-large   & 27.14 &  87.38  & \textbf{3.81}&  \textbf{3.53}  \\
 \hline
 \end{tabular}
 }
 \caption{Generating changes resulting from a complex event (Task 3). See appendix \cref{app:gen-changes-from-action} for the full results. The best scores are bolded.}
 \label{tab:gen-changes-from-action}
 \vspace{-3mm}
 \end{table}
\subsection{Task 4: Identifying Types of Changes}

We fine-tune base models on a multi-label (n=5) binary classification task and assign change mode labels, $(c_1, c_2,..,c_k)$, for an input context consisting of two sentences: story $S$, and $(PI, c, ED)$ where c is a connector phrase. The value of $c$ is ``caused this:'' when $PR$=``agent,'' and ``experienced this:'' when $PR$=``patient.'' %
The results from this classification are reported in \cref{tab:identify-post-conditions} and consist of Subset accuracy, Hamming Score and Macro F1. The \textit{Enc Only}  models consists of a T5 encoder model with a classification head on top.  The classification head consists of the following sequence of transformations: Dropout (p = 0.3) -->~Linear(768x 512) --> tanh() --> Dropout (p = 0.3) --> Linear(512 x 5) --> sigmoid().  We also fine-tuned a pretrained T5 encoder-decoder model in a text-to-text multi-label RTE setting.   

\noindent\textbf{Discussion:} 
These results indicate that while the fine-tuned models are good at generating change summaries, assigning the various change model labels is a challenging task for these LMs.

\begin{table}[t]
\centering
\resizebox{.98\columnwidth}{!}{ \begin{tabular}{|l|c|c|c|}
\hline
Model  &  \emph{Subset Acc } &  \emph{Hamming Score}  & \emph{macro F1}   \\  
\hline
Bart-base  &  64.6 & \textbf{71.7}& 61.2\\
T5-base (Enc Only) & 59.4&  66.1&50.3  \\ 
T5-base (Enc-Dec)& \textbf{65.8}  & 67.3& \textbf{62.0 }\\ 

\hline
\end{tabular}
}
\caption{Results for identifying various Change Modes in Participants (Post Conditions) resulting from the endpoint of a complex action (Task 4).  }\label{tab:identify-post-conditions}
 \vspace{-4mm}
\end{table}
 
\subsection{Task 5: Assessing Participant Involvement}
We turn the 5-point Likert scale for $PI$ into a binary class: the first two options (unlikely to be involved) make up the negative class and the latter three (neutral to likely to be involved) are the positive class. %
We formulate participant involvement and enablement of changes as entailment: the story $S$ is the premise and the hypothesis is framed as the $P$'s involvement in the changes of state indicated by $CS$.\footnote{%
We formulate agent-based hypotheses as ``What the actions of $P$ caused was this: $CS$,'' and patient-based hypotheses as ``What happened to $P$ was this: $CS$.''%
} %
We fine-tuned models on all story annotations; only annotations where $P$'s semantic role is ``agent''; and only the annotations where $P$'s semantic role is ``patient.''  Results are in \cref{tab:identify-involvement}.  

\noindent\textbf{Discussion:} While all the models are able to classify a participant's involvement and enablement of changes with high accuracy there is still room for improvement.  Error analysis indicated models are not able to identify enablement when there are no state changes. This usually happens when the complex action is a hypothetical situation or the changes involve subtle cognition (discussed in Task 3). In T5 models, we noticed some errors contradicted the hypothesis statement; these may be due to the model's external knowledge from pre-training, but this requires further study. 

\begin{table}
\centering
\resizebox{.95\columnwidth}{!}{
\begin{tabular}{|l|c|c|c|}
\hline

 & Combined  & Agent  & Patient  \\

Model  &  Acc./F1 &  Acc./F1 &  Acc./F1  \\
  
\hline
Bart-base & 82.7/76.8 & \textbf{80.7/ 75.2} & 84.2 /76.4 \\
Bart-large & 76.0/43.2 & 75.5/49.1 & 79.2/50.8 \\
T5-base & 82.6/76.5 & 80.1/73.2 & 84.2/77.4 \\
T5-large  &  \textbf{83.0/77.4}  &  80.3/74.3 & \textbf{84.5/78.1}  \\
\hline
\end{tabular}
}
\caption{Identifying Participant's involvement  (Task 5). The best results are bolded. The different folds of the Bart-large models converge at different checkpoints resulting in lower average scores but the best scores for any fold are comparable to the Bart-base model.  
}
\label{tab:identify-involvement}
 \vspace{-4mm}
\end{table}

%% file: sections/ablation_study.tex
\section{Effect of Discourse Text on Models}
\label{sec:ablation-study}

We study the effect of discourse text on model generations of endpoint descriptions using the two story types we annotated: ROCStories and newswrire stories. ROCStories are simpler with short, concise and focused salient events, while newswire are more complex, containing more text not always salient to the complex action we annotated. 
We wish to answer the following questions: 
\begin{enumerate*}[(1)]
\item How does training domain affect endpoint generation? 
\item Are the endpoint generations more/less concise, varied, focused and factual for the story context? 
\item Do models trained on one type of stories transfer their learnt knowledge to generate equally good endpoints for the other type?
\end{enumerate*}

We fine-tuned BART and T5 base models separately on ROCStories vs. Newswire, and evaluated them on test sets for both story types. 
We calculated human scores from the endpoint evaluation HIT. %
From \cref{tab:gen-roc-versus-newswire}, 
we observe the following:  
\begin{enumerate*}[(1)]
\item ROCStories models generate shorter, more varied and abstract descriptions. 
\item Newswire generations are longer and more extractive.
\item ROC-trained BART has significantly lower salience when tested on Newswire stories. News-trained BART does not suffer from poorer salience. %
News-trained T5 has lower salience when tested on ROCStories, while ROC-trained T5 does not result in significantly lower salience.
\item BART generations are less factual than T5, possibly because of higher abstractness~\cite{DBLP:journals/corr/abs-2108-02859}.
\item ROC-trained T5 and News-trained BART obtain similar high scores for factuality and salience.  
\end{enumerate*}

 \begin{table}
 \centering
 \resizebox{.98\columnwidth}{!}{
 \begin{tabular}{l|l|l|c|c|c|c|}
 \cline{2-7}
 Train-on & Model & Test-on& Len & Ext$\downarrow$ & Fact. & Sal.\\
 \hline
 \multirow{4}{*}{ROC} & Bart-base & ROC  & 5.6 & .55 & 4.43 & 4.25 \\
& T5-base & ROC  & 8.4 & .55 &  4.43 &  4.24 \\
& Bart-base & News & 9.8 & .39 & 3.98 & 3.59\\
& T5-base & News & 23.8 & .64 & 4.46 & 4.01    \\
 \hline
 \hline
\multirow{4}{*}{News} & Bart-base & ROC  & 6.2 & .89 & 4.55 & 4.09 \\
& T5-base & ROC   & 9.0 &.74 &   4.45 &  3.43 \\
& Bart-base & News & 11.5 & .77 &   4.62 & 4.01\\
& T5-base & News & 15.5 &.75 &  4.59 & 3.95    \\
 \cline{2-7}
 \end{tabular}
 }
 \caption{Comparing Endpoint generations of models trained on ROCStories and 
 on Newswire stories.  See appendix \cref{app:gen-roc-versus-newswire} for the full results. }

\label{tab:gen-roc-versus-newswire}
 \vspace{-4mm}
\end{table}

%% file: sections/conclusions.tex
\section{Conclusions}
\label{sec:conclusions}

We have argued that a deeper understanding of complex events can be achieved by examining their cumulative outcomes, grounded as changes of state. %
By focusing on a specific participant in a complex event, and a broad notion of its semantic role, we developed a crowdsourcing protocol to obtain 7.7k annotations about complex events and participant state change across 4k stories. We validated 20\% of the annotations, with  high inter-annotator agreement. We have formulated five challenge tasks that stress model's understanding of story outcomes, state changes and complex event understanding. Our evaluations suggest that additional modeling advances are needed to achieve this understanding; we hope that our dataset spurs this future work.

%% file: sections/limitations.tex
\section{Limitations}
\label{sec:limitations}
We acknowledge the following limitations of our approach:
\begin{itemize}[leftmargin=*,itemsep=0pt]
    \item Though the documents we base our annotations on come from well known data sources, our efforts focus on more formal levels of written English. Generation and classification abilities can vary as the formality, style, or language change.
    
    \item Though our work is heavily grounded in inter-disciplinary literature, we adopt a limited two-argument view of complex event participants: either they are an ``agent'' or a ``patient.'' Expanding to other types, or finer-grained notions, of arguments requires more investigation.
    
    \item We use large, pre-trained language models in our experiments. While powerful, they can echo biases, either implicitly or explicitly. We do not attempt to control for these in this work.
\end{itemize}

%% file: sections/appendix-HIT.tex
\section{Additional Details on Data  Preparation}
\label{app:dataset:prep}
In this section, we expand on data processing described in \cref{sec:story-prep}.

\paragraph{Document Selection}
From the Annotated New York Times (ANYT) newswire  articles, we found that stories from the Financial, National and Foreign desks contained the type of complex events that were most reliable to annotate: those with focused discourse text that required less external, societal, or cultural knowledge to understand the story. We did not specifically target obituaries as they could lead to less varied endpoint and cumulative state changes. %
The ROCStories were randomly sampled, and we subsampled stories from ESTER that had \texttt{subevent} annotations in that dataset. %

\paragraph{Participant Identification}
We used spanBERT \cite{mandar_spanbert} to resolve coreferent mentions in the text, and selected the largest cluster of mentions.  To  find clusters containing a valid participant, we selected the shortest text span from all the mentions in the cluster making sure that it is atleast 3 characters long and matched it with the names database published by the SSA.   This ensured that the "participant or prop" we selected is a person, place, group or organization.  In ROCStories and ESTER, the largest cluster is always a person, place or group and did not require this name filtering.

\paragraph{Continuant Story Lines}
We selected the first few lines containing approximately 100 tokens, which resulted in stories similar in length to previous work \cite{han2021ester,glavas-etal-2014-hieve,ogorman_inproceedings}.  We highlighted mentions of the participant to outline a ``continuant'' storyline, i.e., a set of related events that lead to a coherent story involving the participant.  Focusing on the events in a continuant story line helps an annotator observe a complex action and its effects.  By identifying and highlighting a single participant we limit the scope of possible valid  endpoints an annotator might consider.  Assigning a semantic role to the participant, of an agent, or a patient, helps cue the annotator to identify a participant's role in the complex action, and the changes resulting from it.

\section{Additional Details on Crowd Annotation}
\label{app:HIT_information}

\subsection{Worker Qualifications}
\label{app:HIT_information:qualifications}
We did not use requester generated qualification tests to filter out workers because we target the understanding of everyday reported events, not domain or expert-level matter. However, we used community standard quality criteria, such as requiring a 98\% or greater HIT acceptance rate and the completion of 1000 approved HITs. In addition, we required the worker's stated location to be in the USA, UK, Canada, Australia, or New Zealand. Given the language-dependent semantic phenomena we pursue in this work, this location requirement was used for avoiding language-based artifacts.   While qualification tests can filter for spam, initial misunderstandings could exclude capable workers who benefit from additional feedback.  By providing positive and constructive feedback to ensure workers understood the task, we were able to retain workers who improved over time and provided quality annotations, a requirement for any crowdsourced task.  %
  
\subsection{Annotation HIT Streamlining}
\label{app:HIT_information:streamlining}
 Our initial development tested selection of textspans vs. free form text and noticed workers preferred one over the other for some of the steps.  To reduce annotation time, we refined the HIT to prime workers to hone in on the salient information in the story, provided functionality that allowed for a quick highlight and paste of relevant text when needed.  We encouraged free form text in steps 1 and 2., an easy to fill in template for step 3, and highlight and paste for step 4.   
 
 Despite instructions to be concise, early annotations suggested that some workers would try to include as much information as possible into the free form text fields, resulting in lengthy descriptions that provided too much detail (e.g., going beyond immediate outcomes, or providing explanation/justification for why those changes happened). %
To address this issue, we implemented two-tiered length limitations on the free form text. The first tier was a ``soft constraint'': if, e.g., a worker typed in a endpoint greater than 8 words, they were prompted to consider revising, but they did not have to. The second tier was a ``hard constraint'': if, e..g, the endpoint was greater than 15 words, they were prevented from submitting until they rephrased and satisfied the hard constraint limit. %
These limits were set based on the examination of early annotations.

In addition, in each HIT batch, we included a mix of the lengthier Newswire and short ROCStory texts to reduce the monotony of annotation.  From the alpha run annotation times, and internal annotation timing, we estimated the average annotation time for a HIT completion to be under 2 minutes. The bulk of annotations for our HITs were completed within 5 minutes, with a median and mean of approximately 2.5 minutes.  

\subsection{Worker Training and Pay}
\label{app:HIT_information:training-pay}

Our HITs were priced to target an hourly pay of \$10-\$12. We carefully tracked and analyzed the user response times across pilot runs to arrive at the HIT pricing.  For each worker, we carefully examined the first 10-30 annotations to check task understanding, providing feedback and a bonus as appropriate to compensate for the time spent on communication.  We initially had an additional 26 crowd workers who attempted the HIT, but we removed their annotations from the dataset for obvious bad-faith efforts (10 workers) and for benefit-of-the-doubt good faith efforts but where the workers (16 workers) did not follow instructions even with repeated feedback.  Anyone construed to have completed the annotation in good faith was paid, even when their responses were not included in our dataset. 
 
\subsection{Annotation Quality Checking}
\label{app:HIT_information:checking}

Our cursory visual check of the annotation and an automatic lexical check of a list of novel unigrams that are not part of the story ensured the annotation content was focused on the story.  We gave iterative feedback to workers not following task instructions and excluded them with a qualification type when there was no improvement.  

\begin{figure}[t]
\includegraphics[width=1.0\linewidth]{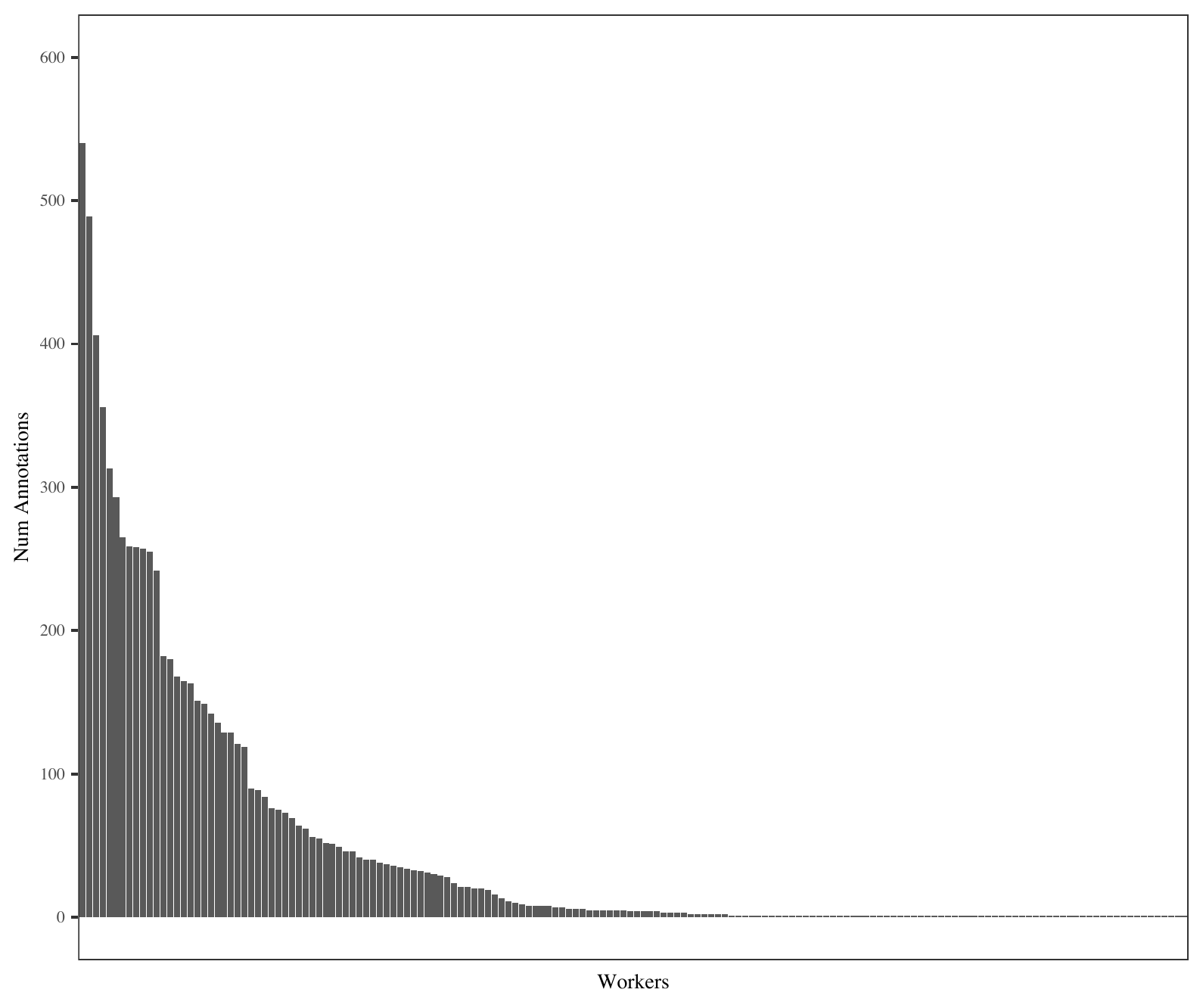}
\caption{Distribution of story annotations completed by workers}
\label{fig:worker-distribution}
\end{figure}

\paragraph{Additional Data Analysis}
In \cref{fig:worker-distribution} we show the distribution of crowd workers for our annotation effort. 

\subsection{Annotation HITs}
\label{app:HIT_information:annotationHITs}
The HIT for acquiring story annotations from crowd workers displays two different views based on participant's semantic role in the story.  General and specific instructions for Step 1-3 are different in two views.

\begin{figure*}
\includegraphics [width=1.0\linewidth] {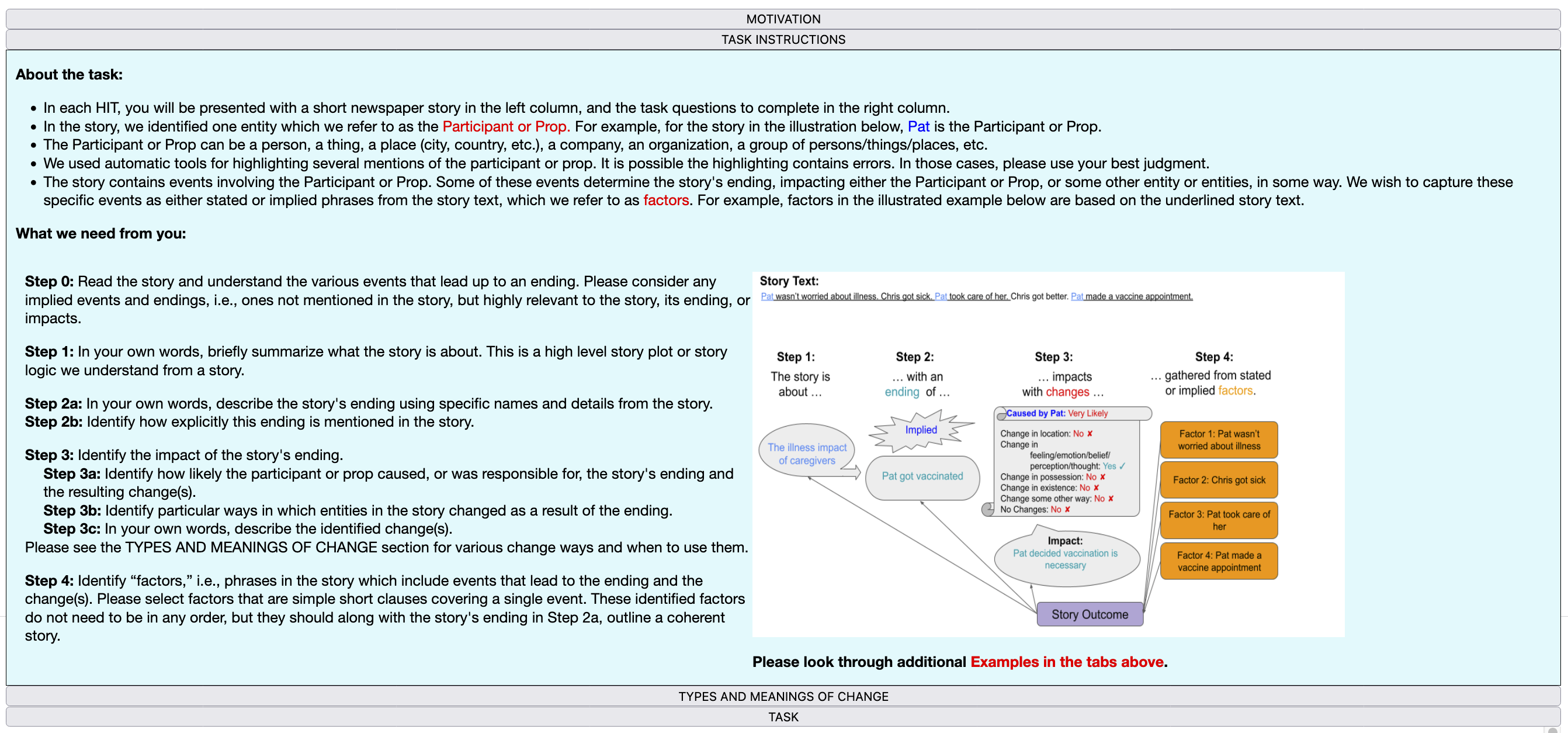}
\caption{Instructions provided for the Agent view of the annotation HIT. The distinguishing aspect that makes it the Agent view is in step 3, where changes are attributable to what the participant or prop caused. In \cref{app:Annotation_HIT_agent_1,app:Annotation_HIT_agent_2,app:Annotation_HIT_agent_3,app:Annotation_HIT_4} we show the interface for each of the steps.} 
\label{app:Annotation_HIT_Agent_Instructions}
\end{figure*}

\begin{figure*}
\includegraphics [width=1.0\linewidth] {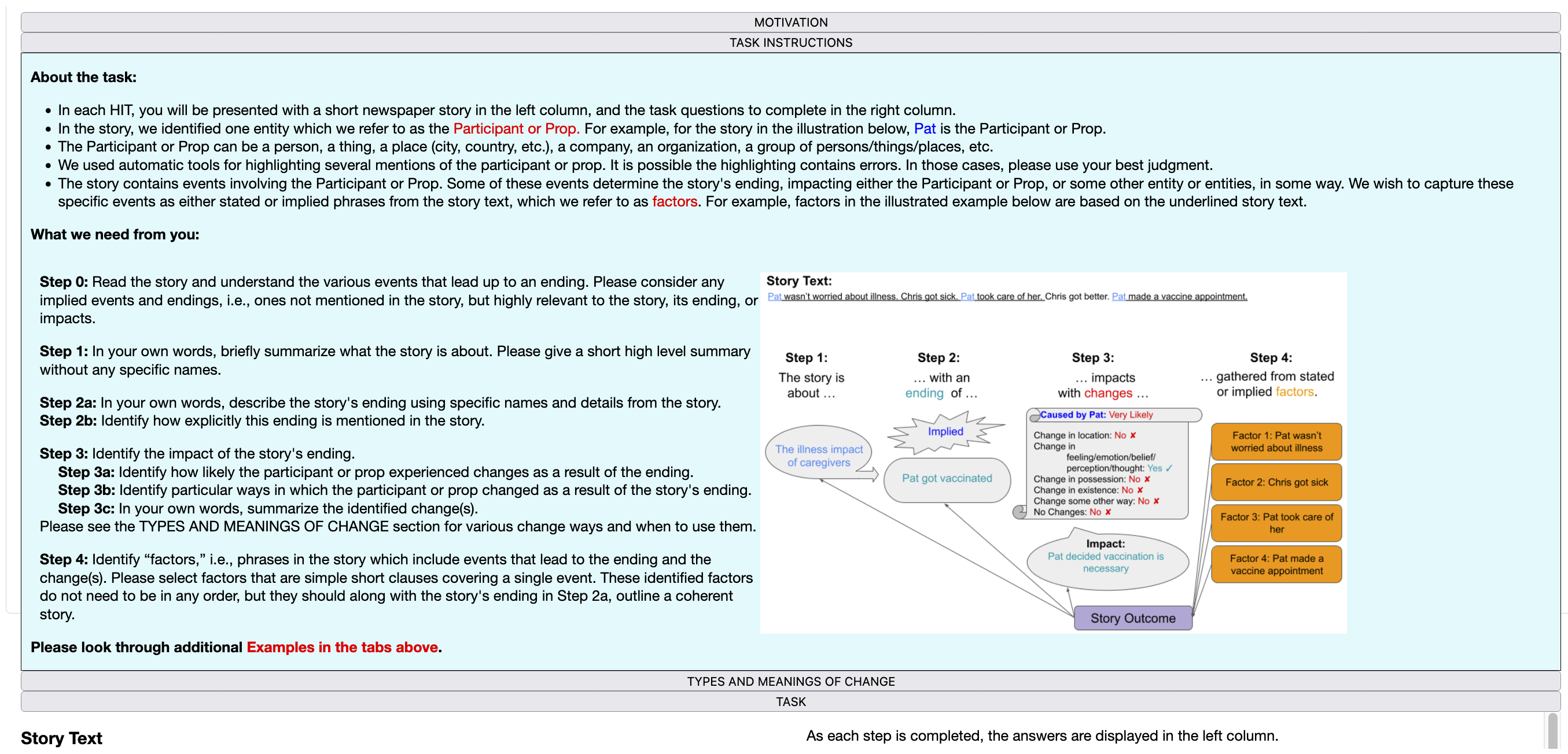}
\caption{Instructions provided for the Patient view of the annotation HIT. The distinguishing aspect that makes it the Patient view is in step 3, where we ask about changes the participant or prop likely experienced. In \cref{app:Annotation_HIT_agent_1,app:Annotation_HIT_patient_2,app:Annotation_HIT_patient_3,app:Annotation_HIT_4} we show the interface for each of the steps.} 
\label{app:Annotation_HIT_Patient_Instructions}
\end{figure*}

\begin{figure*}
\includegraphics [width=1.0\linewidth] {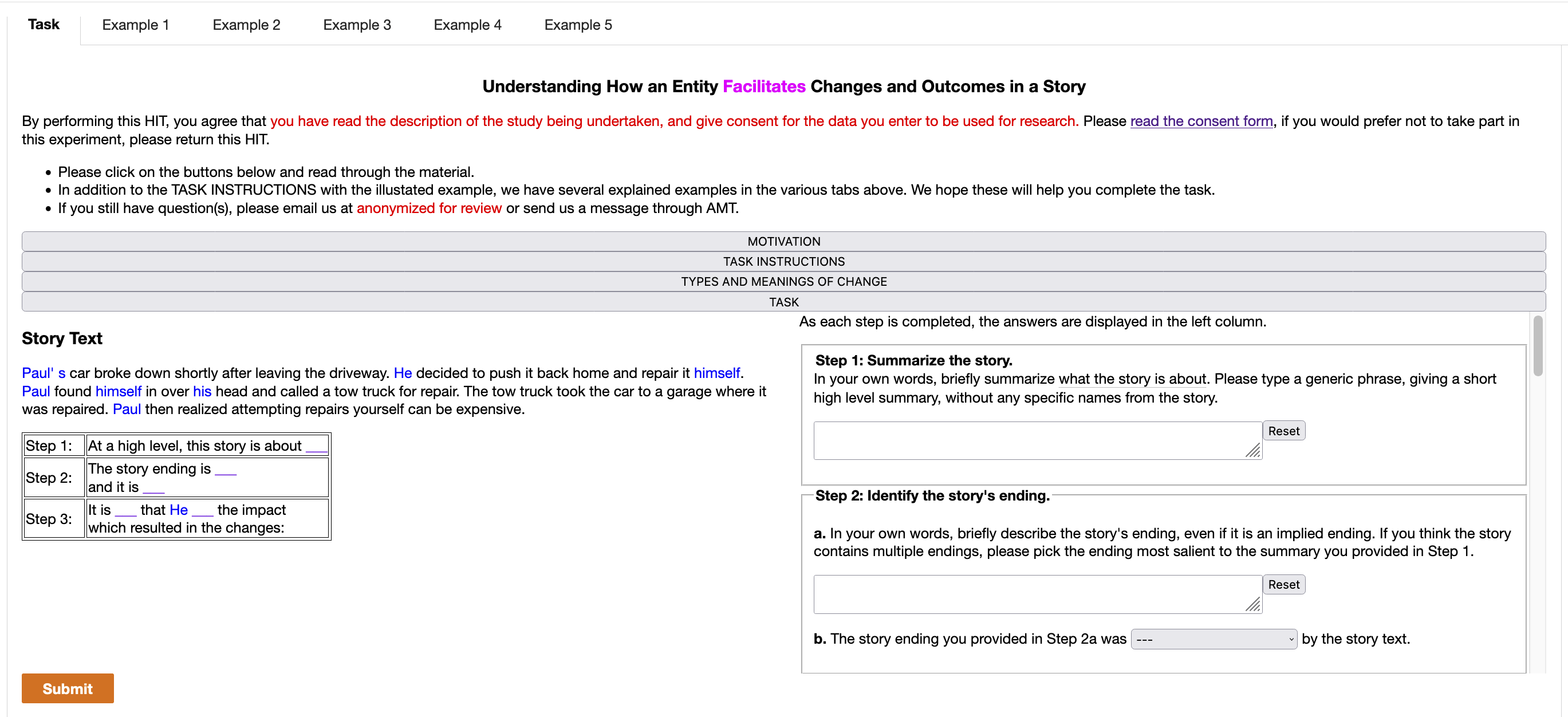}
\caption{Steps 1 and 2 of the Agent view of the annotation HIT.  The Patient view for these steps is similar except for the title of the HIT.} 
\label{app:Annotation_HIT_agent_1}
\end{figure*}

\begin{figure*}
\includegraphics [width=1.0\linewidth] {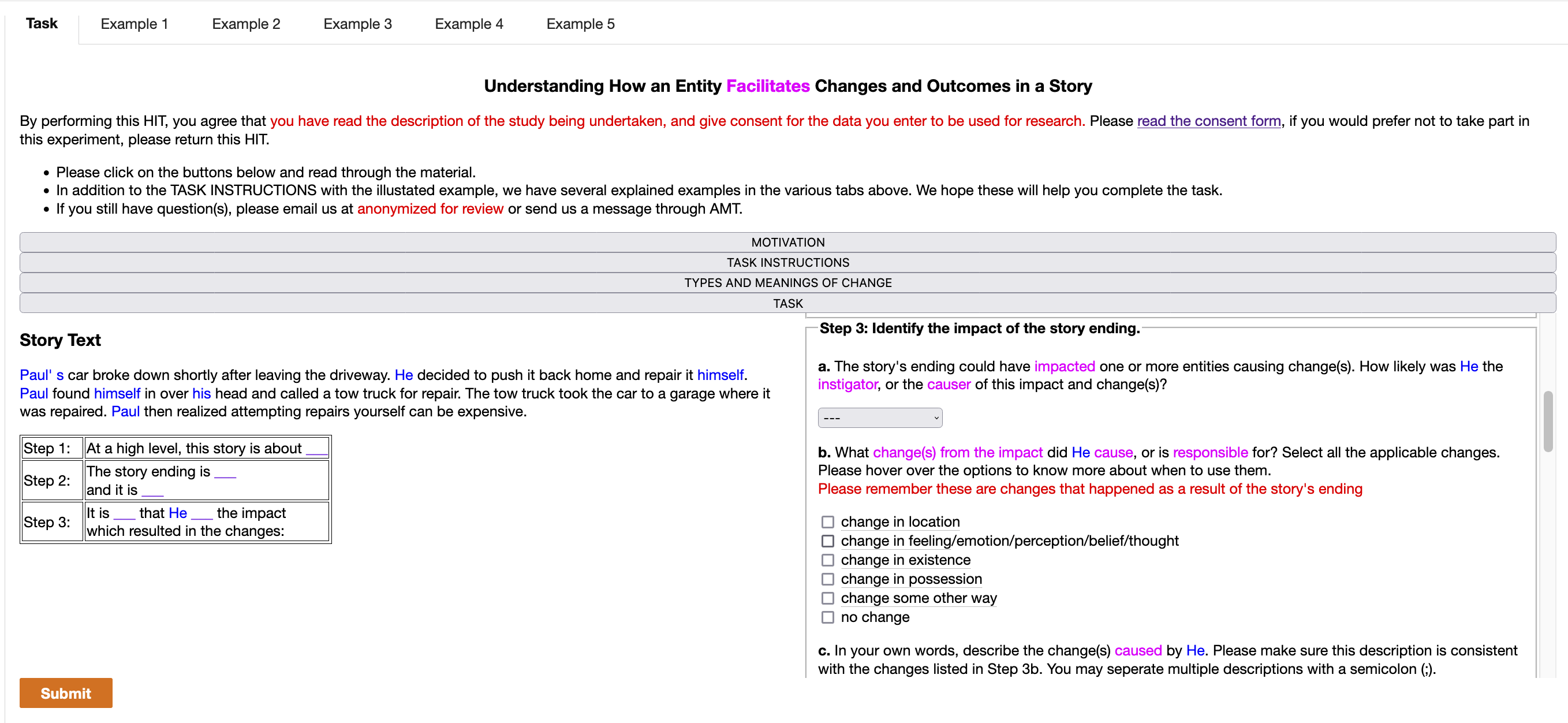}
\caption{Steps 3a \& 3b of the Agent view of the annotation HIT.} 
\label{app:Annotation_HIT_agent_2}
\end{figure*}


\begin{figure*}
\includegraphics [width=1.0\linewidth] {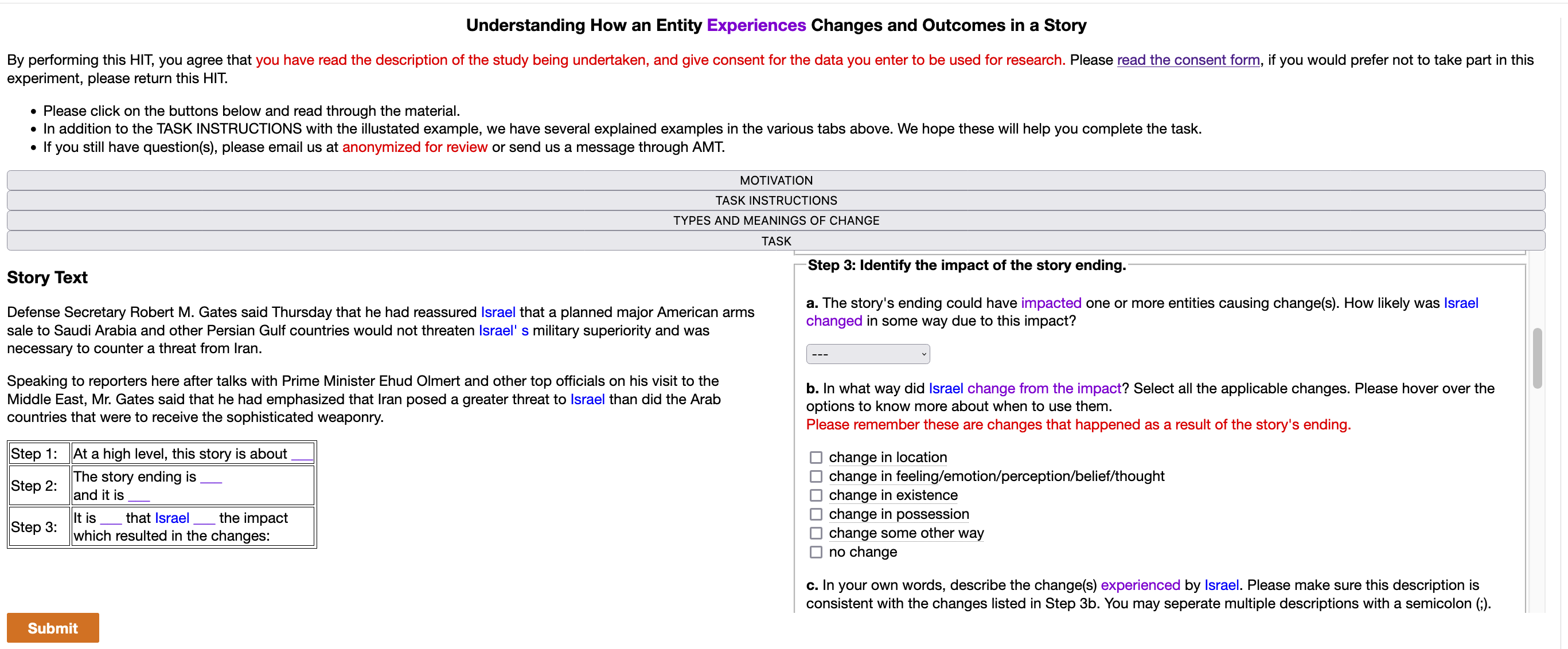}
\caption{Steps 3a \& 3b of the Patient view of the annotation HIT.} 
\label{app:Annotation_HIT_patient_2}
\end{figure*}

\begin{figure*}
\includegraphics [width=1.0\linewidth] {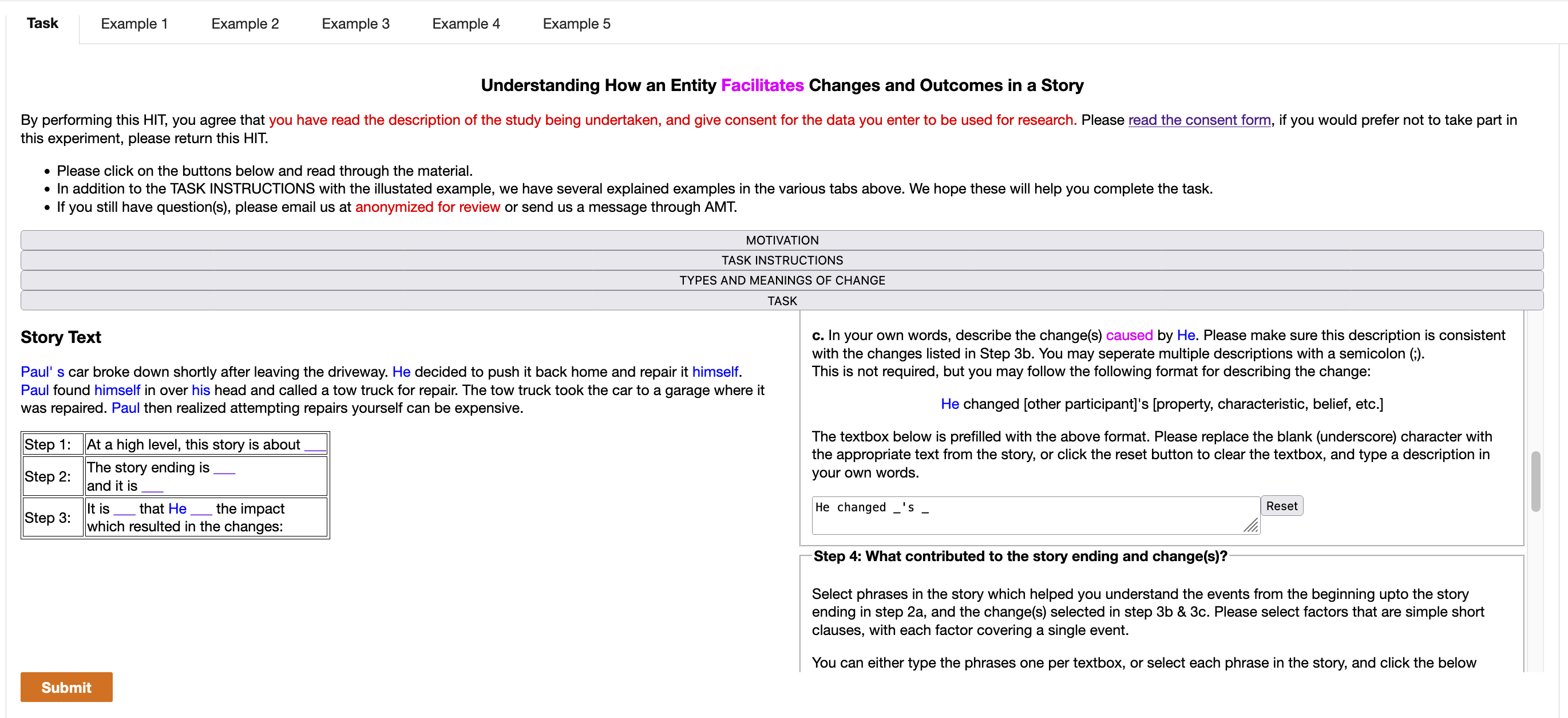}
\caption{Step 3c of the Agent view of the annotation HIT.} 
\label{app:Annotation_HIT_agent_3}
\end{figure*}

\begin{figure*}
\includegraphics [width=1.0\linewidth] {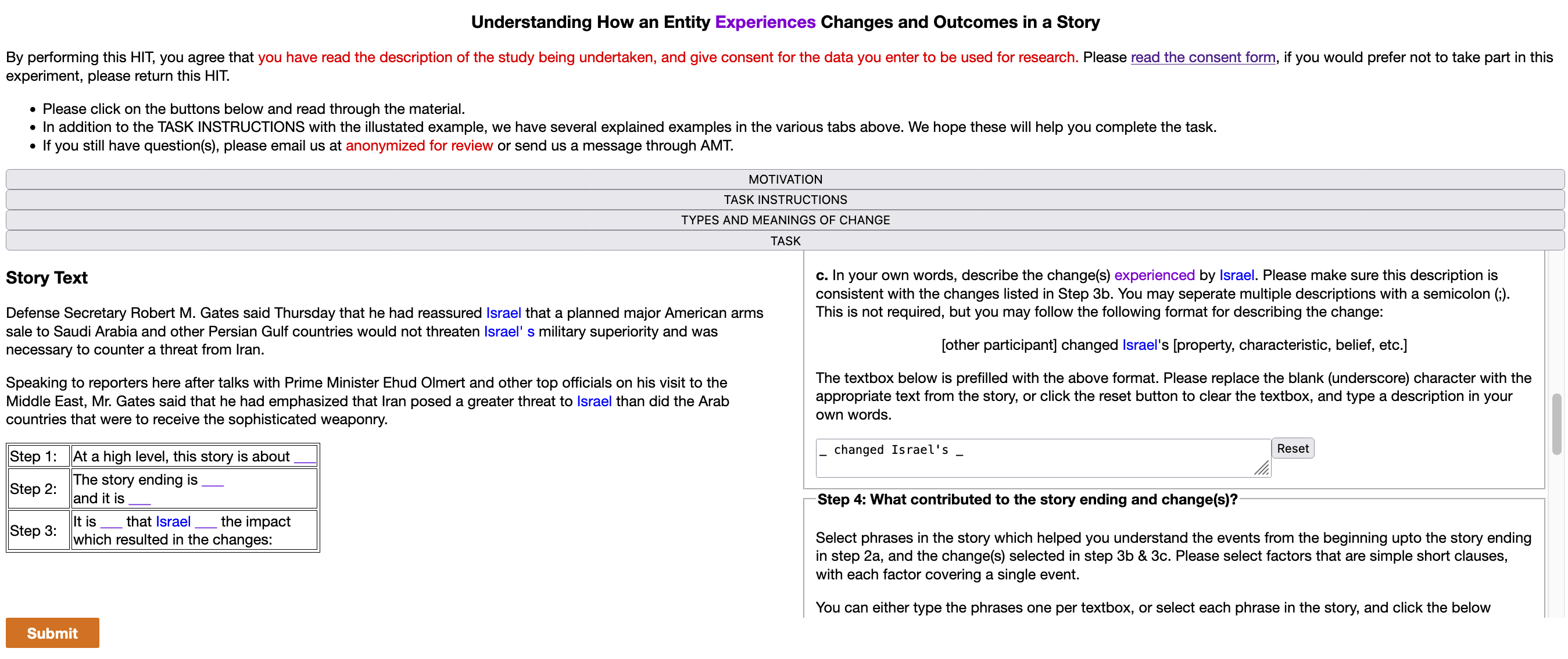}
\caption{Step 3c of the Patient view of the annotation HIT.} 
\label{app:Annotation_HIT_patient_3}
\end{figure*}

\begin{figure*}
\includegraphics [width=1.0\linewidth] {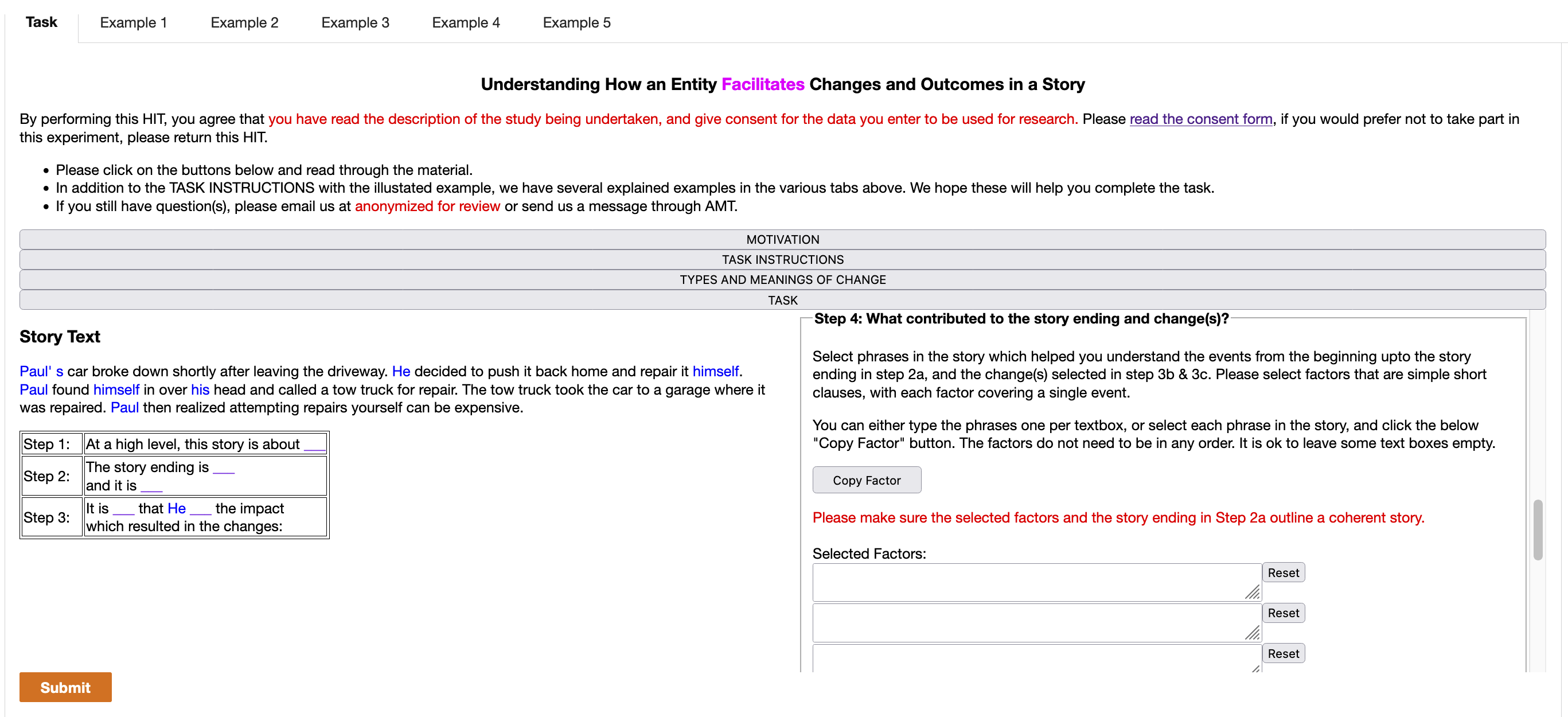}
\caption{Step 4 of the Agent view of the annotation HIT.  The Patient view for this step is similar except for the title of the HIT.} 
\label{app:Annotation_HIT_4}
\end{figure*}

\section{Quality Assessment of Crowd Annotations}
\label{app:annotation-quality-assessment}
In the initial phases of data collection the first two authors of this paper evaluated both the agent and patient views of 50 random stories ($=100$ annotations) to ensure annotator responsiveness and quality. After collecting all the data, 3 crowd workers evaluated a random 1545 annotations and an expert  
evaluated a random 100 of these annotations (equal agent and patient views) for comparison. \cref{tab:annotation-irr} lists the inter-rater reliability (IRR) measured using weighted Fleiss's Kappa~\cite{Marasini2016AssessingTI} with the weighting scheme used by \cite{bastan-etal-2020-authors}, which penalises each dissimilar class by an amount based on the distance between classes (e.g., an item with responses of ``very likely'' and ``very unlikely'' will be penalized more heavily than with responses of ``very likely'' and ``somewhat likely'').   

Our evaluation consisted of 4 validation HITs, where 3 crowd workers rated the various annotation steps (see \cref{app:ValidationHITs}).  The results from this evaluation are listed in \cref{tab:annotation-irr}.  The IRR scores suggest substantial-to-high agreement. %
Notably, these demonstrate that we can obtain high quality process and change of state summaries, endpoint descriptions and enabling sub-event factors.

\begin{table}[t]
\centering
\begin{tabular}{l|r|r|r}
\hline
    \emph{Evaluations} & \emph{Crowd} & \emph{C+E} & \emph{Experts}\\
\hline
1. Process Sum. & 0.81 &  0.81 & 0.90  \\
2. Endpoint Desc. &  0.81 &  0.80 & 0.89  \\
3. Change Sum.  & 0.81 & 0.80 & 0.76     \\
4. Change Modes  & 0.74 & 0.78 & 0.82   \\
5. Factors' Salience  & 0.77 & 0.85 & 0.84   \\
\hline
\end{tabular}
\caption{Inter-rater Agreement scores using weighted Fleiss's Kappa~\cite{Marasini2016AssessingTI}. C+E shows the IRR for the crowd and expert on 100 annotations.  See \cref{app:ValidationHITs} for the various evaluations and what we looked for in the evaluation.}
\label{tab:annotation-irr}
\vspace{-4mm}
\end{table}

\subsection{Evaluation Set Curation}

From the 1545 validated annotations we selected annotations where the average score of the crowd workers for each of the 4 validation HITs is at least 3.0.  This curation resulted in 1196 carefully produced annotations for a given story.  The test data set is made up of these curated annotations and the training data set is made up of the remaining validated and unvalidated annotations.

\subsection{Validation HITs}
\label{app:ValidationHITs}

The various annotation steps are validated using 4 HITs.  3 workers evaluate the following using a 1-5 Likert scale with the options: Strongly Disagree, Somewhat Disagree, Neutral, Somewhat Agree and Strongly Agree. 
\begin{enumerate}[noitemsep]
\item Whether the Process Summary is a valid high level summary of the story. 
\item Whether the endpoint description describes a valid endpoint for the complex action in the story. 
\item Whether the change summary describes changes that happened as result of the complex event described in the story. 
\item Whether the categorization of changes into the five change modes is consistent with the changes inferred from the story. 
\item Whether the factors are salient to the complex event's endpoint. 
\end{enumerate}

\begin{figure*}
\includegraphics [width=1.0\linewidth] {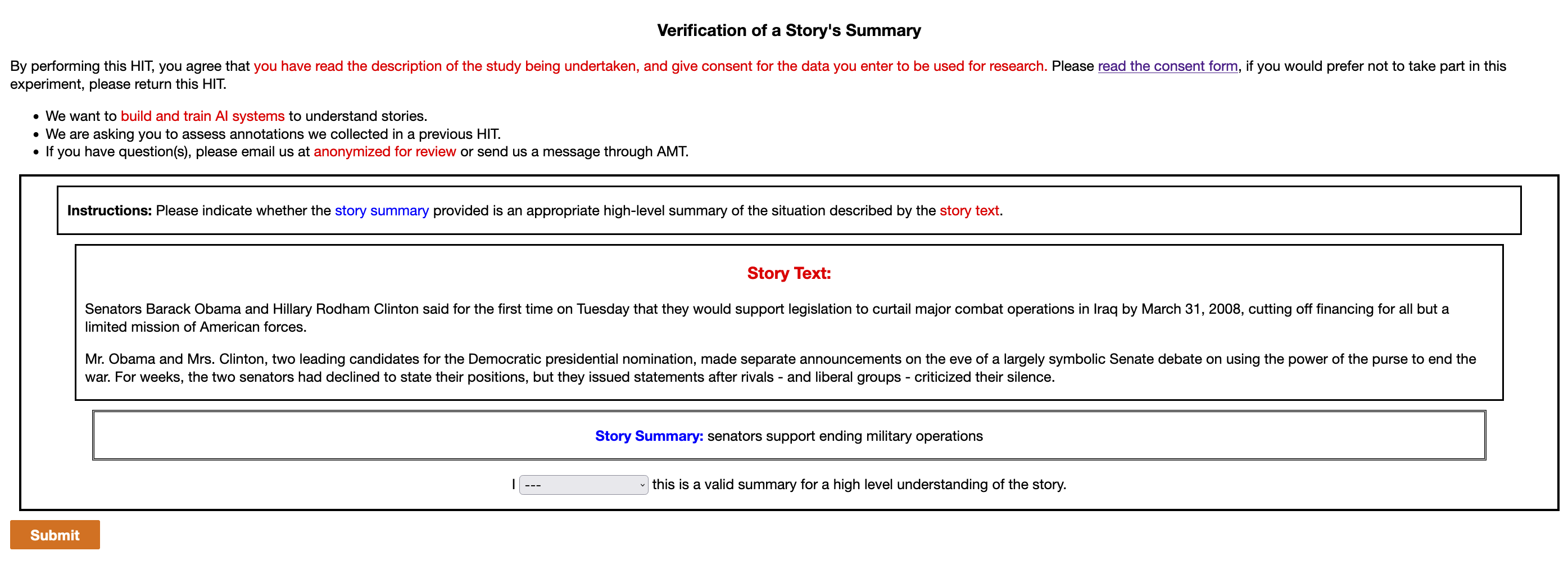}
\caption{HIT for Validating Process Summaries } 
\label{app:validation_process_summary}
\end{figure*}
 
\begin{figure*}
\includegraphics [width=1.0\linewidth] {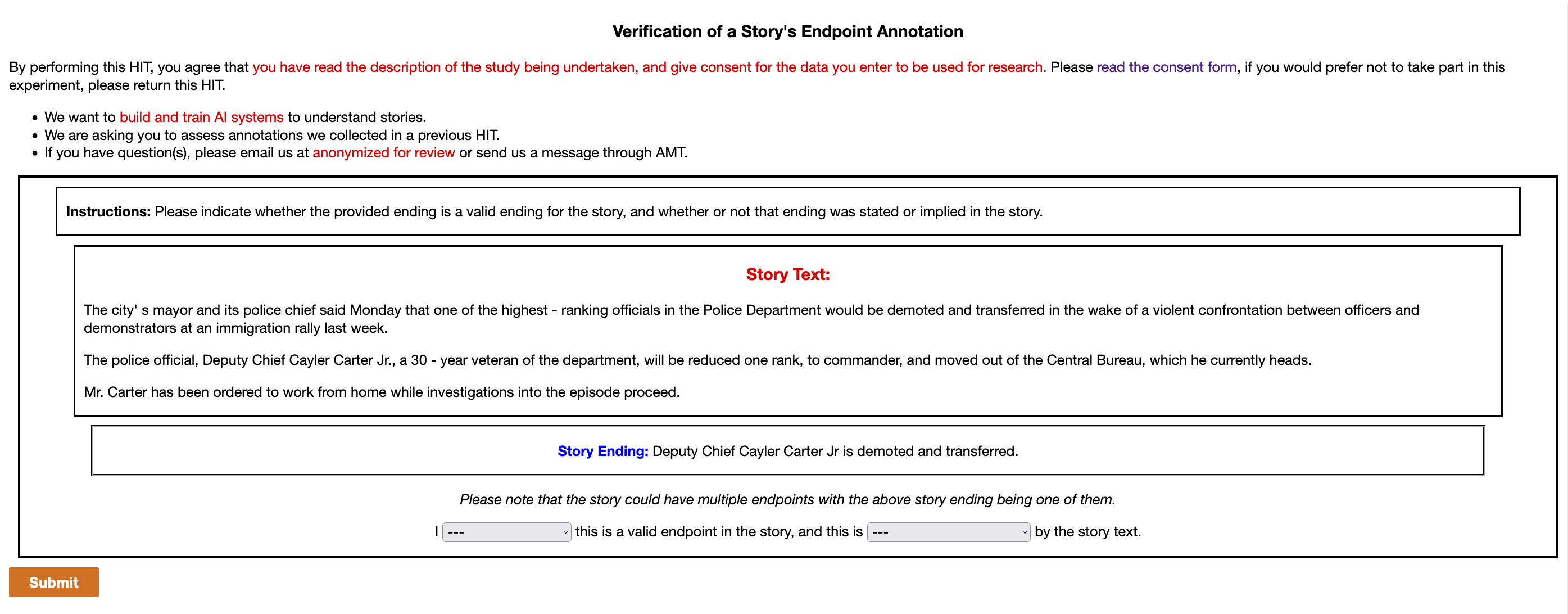}
\caption{HIT for Validating Endpoint Descriptions} 
\label{app:validation_enpoint_summary}
\end{figure*}
 
 \begin{figure*}
\includegraphics [width=1.0\linewidth] {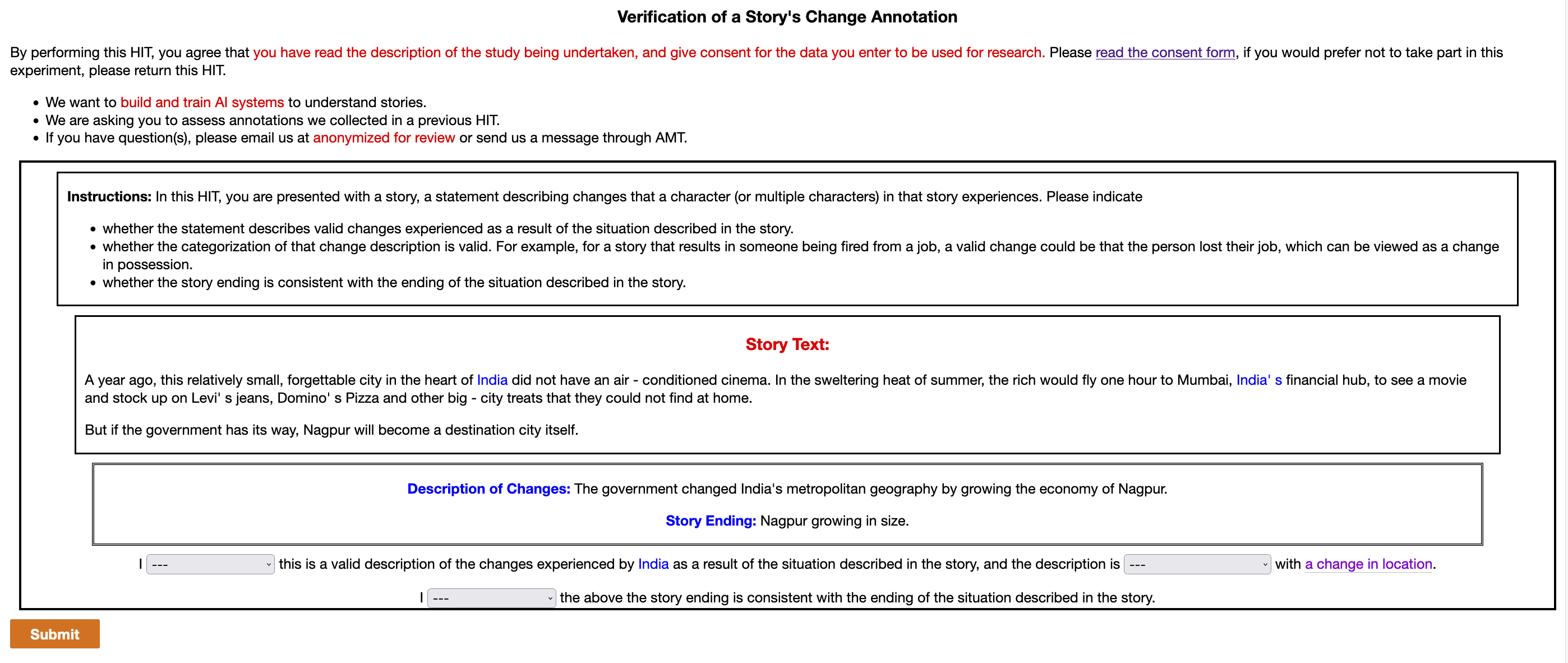}
\caption{HIT for Validating Change Summaries } 
\label{app:validation_change_summary}
\end{figure*}

\begin{figure*}
\includegraphics [width=1.0\linewidth] {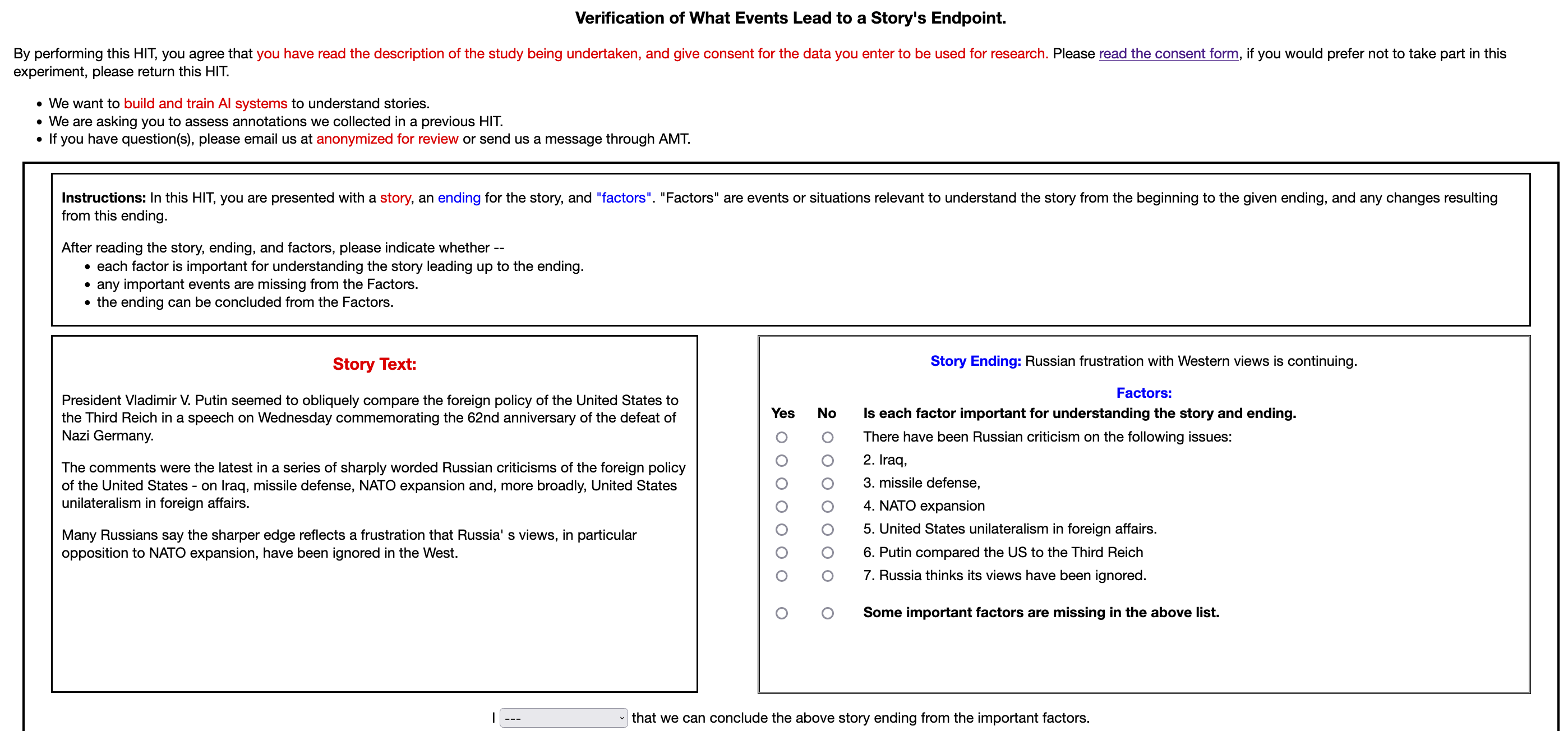}
\caption{HIT for Validating Factors } 
\label{app:validation_factors}
\end{figure*}

\section{HITs for Human Evaluation of Generated Summaries and Factors}
\label{app:human-evaluation-HITs}
\begin{figure*}
\includegraphics [width=1.0\linewidth] {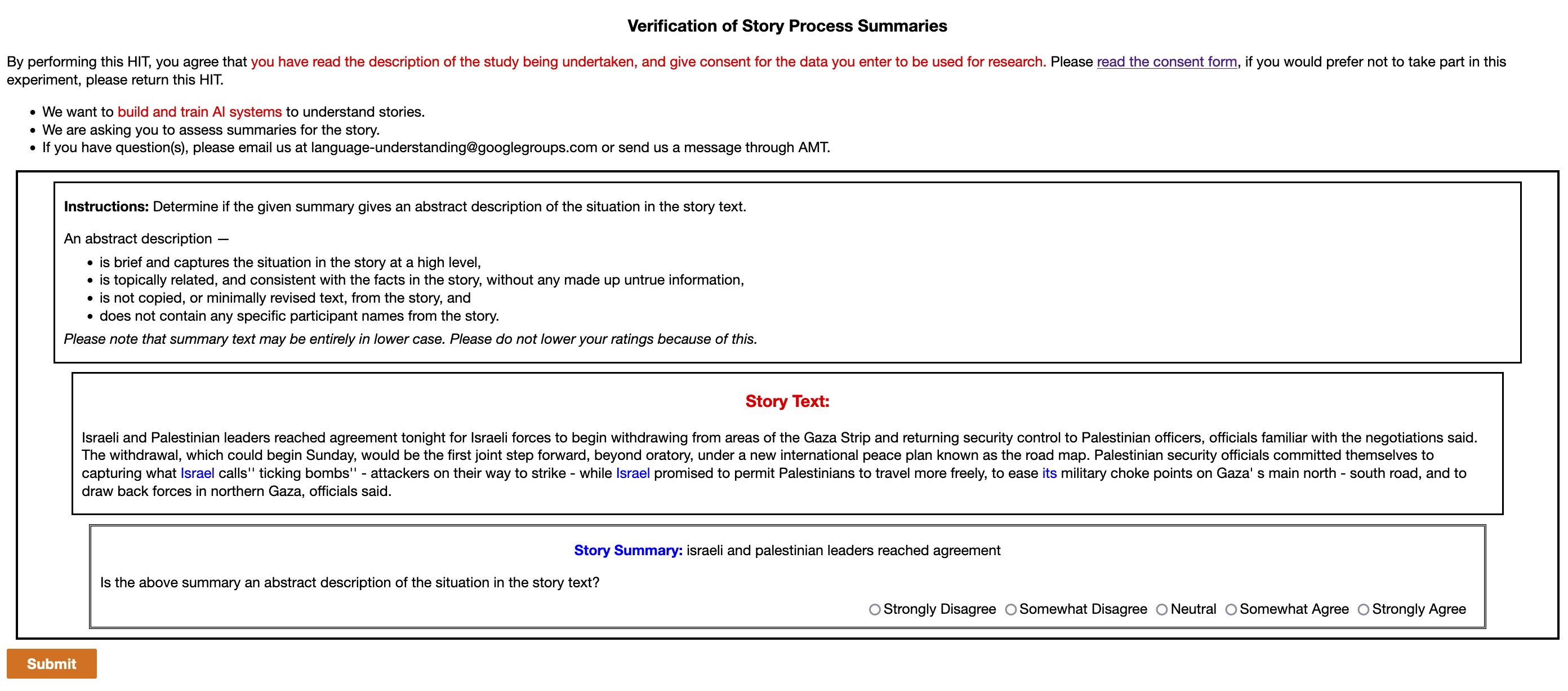}
\caption{HIT for Evaluating Reference, Baseline and Generated Process Summaries } 
\label{app:evaluation_process_summary}
\end{figure*}

\begin{figure*}
\includegraphics [width=1.0\linewidth] {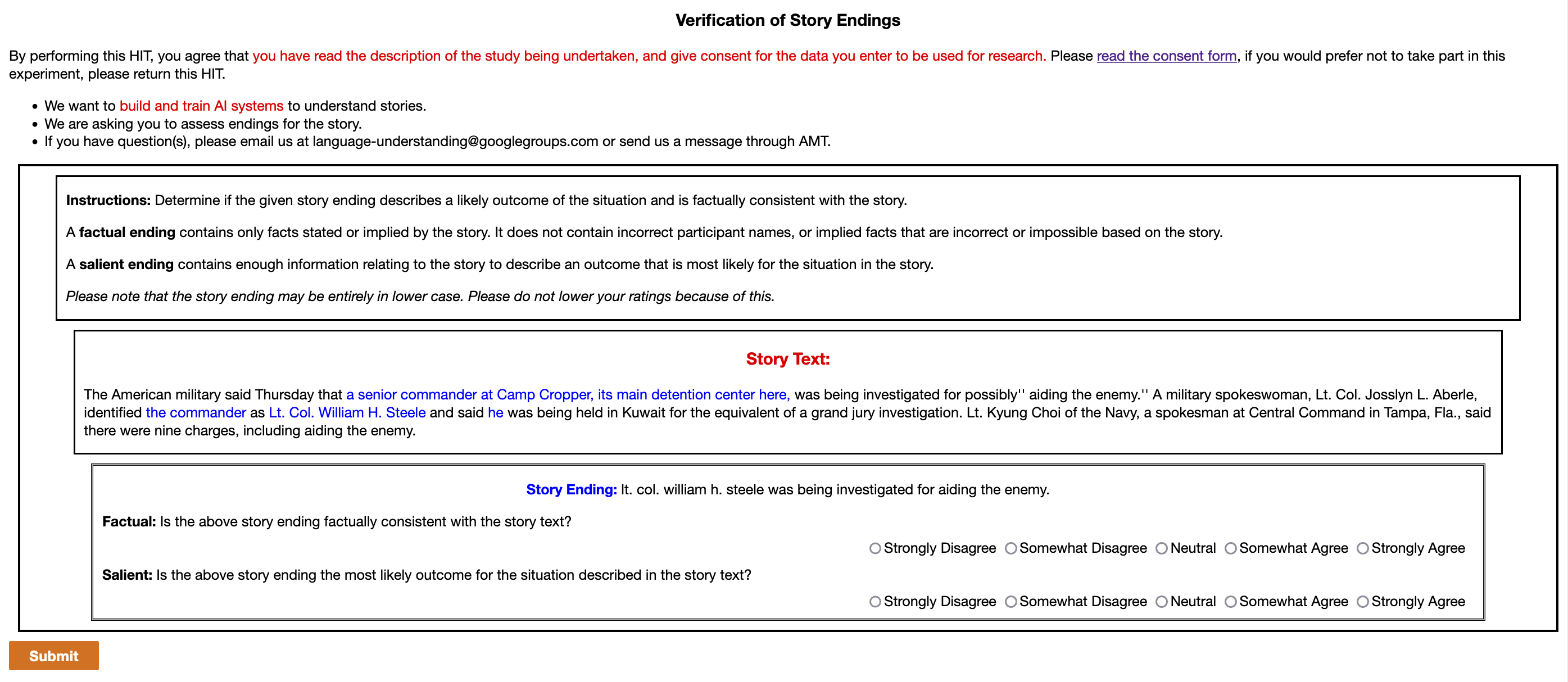}
\caption{HIT for Evaluating Reference, Baseline and Generated Endpoint Summaries } 
\label{app:evaluation_endpoint_summary}
\end{figure*}

\begin{figure*}
\includegraphics [width=1.0\linewidth] {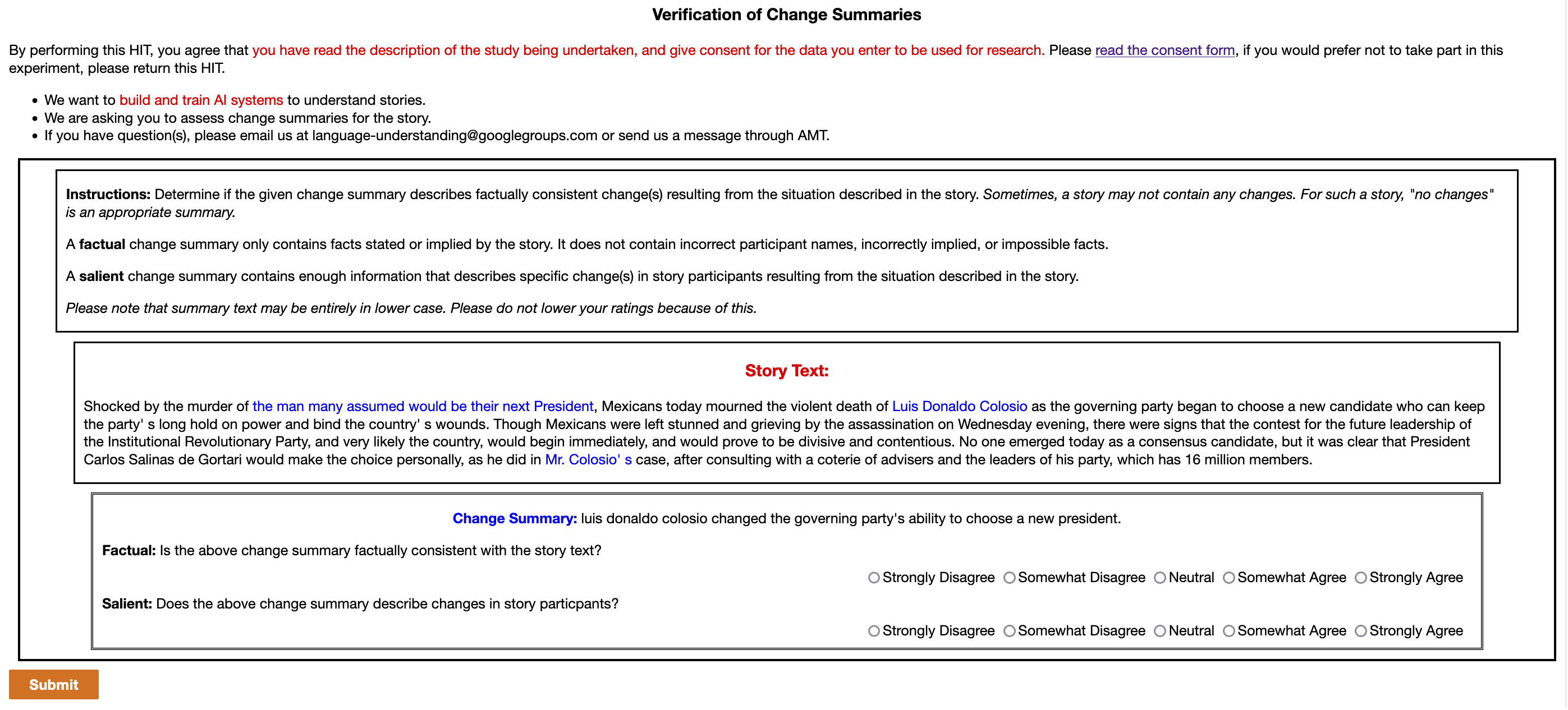}
\caption{HIT for Evaluating Reference and Generated Change Summaries } 
\label{app:evaluation_change_summary}
\end{figure*}

\begin{figure*}
 \includegraphics [width=1.0\linewidth] {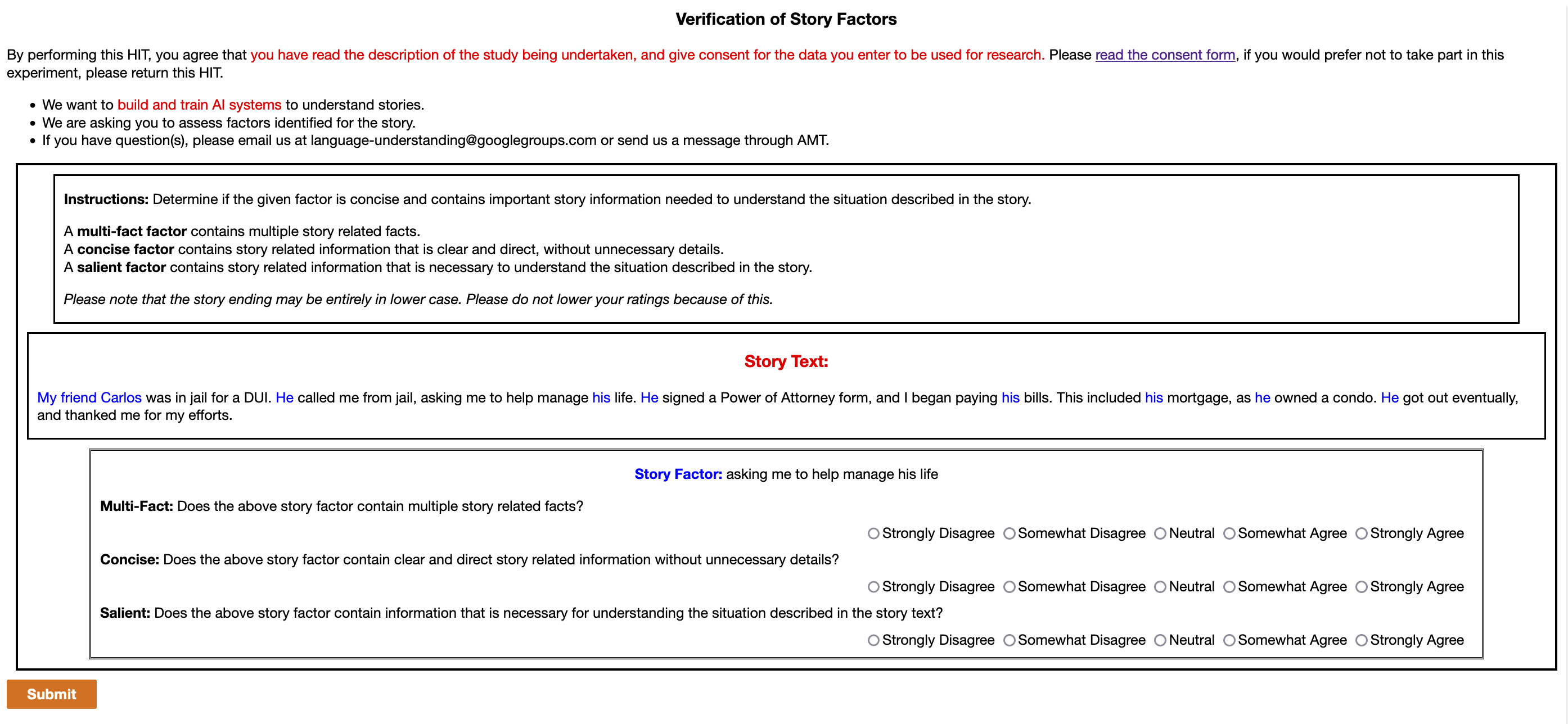}
 \caption{HIT for Evaluating Reference and Generated Factors } 
 \label{app:evaluation_factors}
 \end{figure*}

Here, we show sample UIs for the human evaluation we performed in \cref{sec:human-eval-generation-description} for reference, baseline and model-generated process (\cref{app:evaluation_process_summary}), endpoint (\cref{app:evaluation_endpoint_summary}), change summaries (\cref{app:evaluation_change_summary}) and factors (\cref{app:evaluation_factors}) .  
These HITs are used to evaluate the following aspects of a summary using a 1-5 likert scale with the options Strongly Disagree, Somewhat Disagree, Neutral, Somewhat Agree and Strongly Agree.




1) \emph{Abstractness (Task 1):}  Whether the summary is a brief, high level, abstract description that faithfully captures the complex action in the story.   

2) \emph{Validity (Tasks 2 and 3):}  Whether the summary is a valid ending for the situation described in the story for task 2.   In task 2, when evaluating factors, we check whether each factor contains story related information.  For task 3, we check whether the summary mentions a change that is consistent with the details in the story.  A blank summary  is valid, if the story does not contain any changes.

3) \emph{Salience (Tasks 2 and 3):}  Whether the summary is a valid ending for the situation described in the story for task 2.  In task 2, when evaluating factors, we check whether each factor is necessary for understanding the situation.  For task 3, we check whether the summary mentions a change that is consistent with the details in the story.  When a story does not contain any changes, the annotation of no-changes is considered salient.

%% file: sections/appendix-processing.tex
\section{Model Training}
\label{sec:model-training}
 To perform early development experiments, we used  5-fold cross validation for 2 epochs. We found that while the evaluation loss plateaued after training for an epoch, recall improved with further training for another epoch. Classification results are the average of the 5-fold cross validation after 2 epochs for T5 and 1 epoch for BART for the single label classification in Task 5.   The Multi-label binary classification for Task 4 was trained for 5 epochs for the BART models, 10 epochs for the T5 Encoder Only and the Encoder-Decoder model.  For the summarization tasks, we trained models on the entire training set without subsequent hyperparameter tuning for 2 epochs for all tasks. 
 
\section{Expanded Results}
\label{sec:expanded-results}

In this section we present expanded results from \cref{sec:models}. In addition to average number of tokens in a summary or factor (\textit{Len}) and percentage of extractive trigrams from the story text (\textit{Ext}), ROUGE-L (longest word sequences) and BertScore that were reported in the paper, we use the ROUGE scores based on unigrams (ROUGE-1) and bigrams (ROUGE-2), corpus and sentence level BLEU, and METEOR. %
Unlike the lexical- and ontological-based metrics of ROUGE, BLEU and METEOR, BertScore aims to provide a modern, embedding-based approach for handling semantic equivalence/similarity even when the texts being compared have different surface forms (e.g., different words are used). These automated scores are calculated using all the test data where as the human evaluation scores are for the 50 randomly selected data items for evaluation. The automated scores of ROUGE, BLEU, METEOR and BertScore are not reported for the reference summaries, as these summaries were used as gold references when calculating automated scores for the baseline and model generations. For the human evaluation we report the IAA by crowd workers.  The difference in IAA for the crowd $+$ expert scores and the crowd scores was < 0.1. The metrics reported on the left of the double line are calculated for all the test data.%
The best value for each score category are bolded  and those  that are significantly higher than all other values are marked with a *.   

 \begin{table*}
 \centering
 \resizebox{.98\textwidth}{!}{
 \begin{tabular}{l|l|c|c|c|c|c|c|c|c|c||c|c|c|}
 \cline{2-14}
 &   & && \multicolumn{3}{|c|}{Rouge} & MET- & \multicolumn{2}{|c|}{BLEU} & Bert & \multicolumn{3}{|c|}{Abstractness}\\ 
  & Model&Len  &Ext$\downarrow$& 1 & 2 & L & EOR& sent. & corpus &  Score& AvgLS & IAA&\%Abs\\
 \cline{2-14}
  &Reference &3.6 &.13* & \multicolumn{7}{|c||}{-}&\textbf{3.57$*$} &.82 &72\\ 
 \cline{2-14}
 &About $P$  & 1.7 & .27 &10.55  & 4.22  & 10.43  & 0.07 & 0.35  &2.70&83.86& 2.37 &.48& 16\\
 \hline
 \multirow{3}{*}{Story} &Bart-base  &4.0& .46&22.43 &8.34 &21.43 & 15.26 & \textbf{7.89} & 4.27 & 86.70   & 2.77 &.86& 42\\
& T5-base  &10 &.60  & 21.14 & 7.48 &19.50 & \textbf{18.43} & 5.55&2.81 & 85.99 &    2.32 & .72& 28 \\
& T5-large  &6.9 &.63 & 21.73 & 8.47&20.30 & 17.59 & 7.78 & 4.16 & 86.15  & 2.13 & .77& 20 \\
 \hline
\multirow{2}{*}{Fact.} & Bart-base & 4.2 & .33&\textbf{25.17} & \textbf{9.05}&\textbf{23.81} & 17.61 & 6.73 & \textbf{4.89}&  \textbf{87.66}  &3.22 & .76 & 54\\
& T5-base &9.9 & .56& 19.94 & 6.39&18.29& 17.47 & 4.19 & 2.63&  86.10  & 2.73 & .69 & 36\\
 \cline{2-14}
 \end{tabular}
 }
 \caption{Generating Process Summaries for stories and factors$+$endpoint (Task 1).  The best scores are bolded. * indicates the Reference value for Abstractness is significantly higher than other values in the column with a p value between 0.001 - 0.0001 (except for Bart-base trained on Factors where the p value is 0.13). \textit{Len} value for Reference is considered the best as the baseline value is not meaningful. The column AvgLS is the average of 3 crowd worker 1-5 Likert scores, and the column \% Abs is percentage of instances with a score $\geq 3$.}
 \label{app:generating-actions}
 \end{table*}
 
 \begin{table*}
 \centering
 \resizebox{.98\textwidth}{!}{
 \begin{tabular}{l|l|c|c|c|c|c|c|c|c|c||c|c|c|c|c|c|}
 \cline{2-17}
 &   & && \multicolumn{3}{|c|}{Rouge} & MET- & \multicolumn{2}{|c|}{BLEU} & Bert & \multicolumn{3}{|c|}{Factuality} & \multicolumn{3}{|c|}{Salience}\\ 
  & Model&Tok  &Ext$\downarrow$& 1 & 2 & L & EOR& sent. & corpus &  Score& AvgLS & IAA&\%Abs& AvgLS & IAA&\%Abs\\
 \cline{2-17}
  &Reference &7.9 &.27 & \multicolumn{7}{|c||}{-}&4.15&.87 &80&3.46 &.76 &68\\ 
 \cline{2-17}
 &Last sent.  & 23.3 & .79 &24.08  & 12.33  & 21.82  & \textbf{24.62} & 0.58  &6.59&85.60& 4.49 & .82 &90&3.35 &.82& 62 \\
\hline
\hline
 \multirow{4}{*}{Story} & Bart-base  & 11 & .72 &27.83 & 13.05 & 25.43 & 22.88 & \textbf{11.12} &\textbf{8.31} & 87.61 & 4.66&.73&96&3.97 & .69 &  80\\
&Bart-large & 10.3 & .63&27.00  & 12.51& 24.74& 21.44 & 10.82 & 8.30 &87.62 & 4.59&.78&96&  3.81  & .72  & 74 \\
&T5-base   &13.6 &.70 & 26.66 &11.79& 24.07 & 22.69 & 9.79 &6.62 & 87.19 & \textbf{4.71*}&.75 &96&4.03 &  76& 82  \\
& T5-large  &12.9 &.67 & \textbf{28.36} &\textbf{13.15}& \textbf{25.71} & 24.31 & 10.93 & 7.83 & 87.54 & \textbf{4.71*}&.77&100& \textbf{4.23*} & 72  &90  \\
 \hline
 \hline
\multirow{2}{*}{Fact.} & Bart-base  &7.3 & .49& 26.26& 10.86 & 24.09& 18.28 &9.15&7.32&  \textbf{87.93} &  4.11&.77&80&3.28& .76& 52 \\
& T5-base    &10.4 &.47& 24.74 & 9.52& 22.01 & 18.53 & 7.31&5.69 & 87.07 & 3.99&.72&82&3.02 &69& 44\\
 \cline{2-17}
 \end{tabular}
 }
 \caption{Generating Endpoints for stories and factors (Task 2a).  The best scores are bolded. * indicates the value is significantly higher than the Reference with a p value of .002 for Factuality and .0008 for Salience.   The column AvgLS is the average of 3 crowd worker 1-5 Likert scores, and the column \% Abs is percentage of instances with a score $\geq 3$.}
\label{app:gen-action-events}
\end{table*}

 \begin{table*}
 \centering
 \resizebox{.98\textwidth}{!}{
 \begin{tabular}{l|c|c|c|c|c|c|c|c|c||c|c|c|c|c|c|c|c|c|}
 \hline
     &\multicolumn{2}{|c|}{Factors}& \multicolumn{3}{|c|}{Rouge} & MET- & \multicolumn{2}{|c|}{BLEU} & Bert &\multicolumn{3}{|c|}{Brevity} & \multicolumn{3}{|c|}{Factuality} & \multicolumn{3}{|c|}{Salience}\\ 
 Model&Num &Len  & 1 & 2 & L & EOR& sent. & corpus &  Score& AvgLS & IAA&\%Abs&  AvgLS & IAA&\%Abs&  AvgLS & IAA&\%Abs\\
 \hline
 Reference &3.6 &8.3 & &&&&&&&\textbf{3.35} &.73 &90&3.25&.67 &52&3.04 &.73&42\\ 
 
\hline
\hline
 Bart-base  &3.5&14.0& 49.90& 37.35 & 45.28&  43.04 & 0.42 & 25.18 & 88.06&2.46 & .62 &.60&3.49 & .69 &59& 3.57 & .76&63  \\
 Bart-large  &3.6&13.7& 50.51& 38.11 & 45.98& 43.63 & \textbf{0.44}&25.32 & 88.10 &  2.88 & .65&67 & 3.2 & .68 &50& 3.31 & .73&52\\
T5-base    &2.6&19.6& 47.89 & 35.86& 43.31 & 44.12 &0.38& 24.61 & 87.74 &    1.77 & .66 &38& 3.69 & .76 &68& \textbf{4.01*} & .79 &79\\
 T5-large   &3.7&13.9 & \textbf{52.26} & \textbf{40.57}& \textbf{47.96} & \textbf{46.83} &0.40&\textbf{26.8}&\textbf{88.44}&2.15 & .65 &48& \textbf{3.80*} & .70 &71& 3.96 & .77&78\\
 \hline
 \end{tabular}
 }

 \caption{Generating Factors from stories and their endpoints (Task 2b).  The best scores are bolded. The * for factuality and salience indicates the value is significantly higher than the Reference with a p value of .0001. The Reference value for Brevity is significantly higher than all values in the column with a p value of .0001.   The column AvgLS is the average of 3 crowd workers' 1-5 Likert scores, and the column \% Abs is percentage of instances with a score $\geq 3$.}
\label{app:gen-action-factors}
\end{table*}

\begin{table*}
\centering
\resizebox{.98\textwidth}{!}{ \begin{tabular}{|l|c|c|c|c|c|c|c||c|c|c|c|c|c|}
 \hline
 Model  &  \multicolumn{3}{|c|}{Rouge} & MET- & \multicolumn{2}{|c|}{BLEU} & Bert- & \multicolumn{3}{|c|}{Factuality} & \multicolumn{3}{|c|}{Salience} \\
   & 1 & 2 & L & EOR& sent. &corpus & Score&  AvgLS & IAA&\%Abs& AvgLS & IAA&\%Abs \\

 \hline
  Reference&\multicolumn{7}{|c||}{-} &3.36&.78&60& 3.32 &.76& 64\\   \hline
\hline
Bart-base  & \textbf{39.56} & \textbf{23.66} & \textbf{34.79} & \textbf{28.68} &\textbf{7.97}&5.29 & \textbf{88.39} &3.03 &.69& 48& 2.93 & .71&50 \\
 Bart-large & 39.28 & 20.68  & 32.80 & 27.87 &7.10& 6.91& 88.23 &  2.99 & .72&44&3.05&.74  & 48\\
 T5-base  &  32.82& 14.74 & 26.81 & 24.62 & 5.91&6.73  & 87.20  & 3.74 &.75&52& 3.23&.66 &66\\
 T5-large   & 34.40&  15.53 & 27.14 & 24.65  &6.11& \textbf{7.07}  & 87.38  & \textbf{3.81}&.70&70&  \textbf{3.53}&.70 &72  \\
 \hline
 \end{tabular}
 }
 \caption{Generating changes resulting from a complex event (Task 3).  The best scores are bolded. The column AvgLS is the average of 3 crowd workers' 1-5 Likert scores, and the column \% Abs is percentage of instances with a score $\geq 3$.}
 \label{app:gen-changes-from-action}
 \end{table*}

 \begin{table*}
 \centering
 \resizebox{.98\textwidth}{!}{
 \begin{tabular}{l|l|l|c|c||c|c|c|c|c|c|c||c|c|}
 \cline{2-14}
 Trained- &  & &&& \multicolumn{3}{|c|}{Rouge} & MET- & \multicolumn{2}{|c|}{BLEU} & Bert & {Factu-} & Sali-\\ 
  on & Model & Test-on&Len  &Ext& 1 & 2 & L & EOR& Sent. & corpus &  Score& uality & ence\\
 \hline
 \multirow{4}{*}{ROC} & Bart-base & ROC  & 5.6 & .55 &42.54 & 21.13 & 39.16 & 31.20 & 18.54  & 12.37 & 91.46 & 4.43 & 4.25 \\
&T5-base & ROC   &8.4&.55 & 43.13 &20.66& 38.53 & 33.35 & 16.56 &9.77 & 90.83 &  4.43 &  4.24 \\
&Bart-base & News & 9.8 & .39&22.53  & 8.25& 20.49& 16.27 & 6.46 & 4.97 &87.17 &   3.98  & 3.59\\
& T5-base & News &23.8 &.64& 23.60 &10.03& 20.60 & 24.54 & 6.99 & 4.64 & 86.44 & 4.46 & 4.01    \\
 \hline
 \hline
\multirow{4}{*}{News} & Bart-base & ROC  & 6.2 & .89 &39.39 & 18.80 & 36.41 & 29.64  &15.55 & 10.95 & 90.46& 4.55 & 4.09 \\
&T5-base & ROC   &9.0 &.74 & 35.42 &14.41& 31.39 & 26.94 & 11.75 &7.01 & 90.26 &  4.45 &  3.43 \\
&Bart-base & News & 11.5 & .77&25.28  & 11.58& 23.12& 20.77 &  10.16 &7.78 &86.95 &   4.62  & 4.01\\
& T5-base & News &15.5 &.75 & 22.94 &9.69& 20.51 & 20.22 & 8.05 & 5.48 & 86.51 & 4.59 & 3.95    \\
 \cline{2-14}
 \end{tabular}
 }
 \caption{Comparing Endpoint generations of models trained on ROCStories and models trained on Newswire stories}
\label{app:gen-roc-versus-newswire}
\end{table*}
\section{Computing Infrastructure \& Processing information}

\paragraph{Infrastructure}
We trained our models on a single RTX 8000 with 48GB of GPU memory. Approximate run time was 1 hour.

\paragraph{Model Parameters}
In addition to the number of parameters in each of the models we consider (e.g., Bart-Base, T5-base, T5-large), each of our fine-tuned classification models has a separate classifier layer. This layer takes in a $D$ dimension embedding from the encoder and uses a single linear layer to compute $K$ dimensional logits (therefore, an additional $D x K$ parameters). The value of $D$ will depend on the model, and the value of $K$ will depend on the number of label types that could be predicted. For the generation tasks, we do not add any additional parameters to the models.

\paragraph{Hyperparameters}
For all experiments we used AdamW~\cite{DBLP:journals/corr/abs-1711-05101} optimizer, a learning rate of $10-4$, a weight decay of $10^-4$ and a random seed of 11.   
We applied manual tuning and tried various learning rates from .001 to .00001 as suggested for  the BART and T5 models.  For the generation we used  Top-K sampling with a beam size of 2.  These parameters  worked well for all the models and this was selected based on accuracy for the classification tasks and Rouge scores for the summarization tasks. 

\paragraph{Results statistics}
With our 5-fold cross-validation, the automated metrics for summarization and the accuracy/F1 values for classification varied by less than 3 percent.